\documentclass[preprint,12pt,number]{elsarticle}

\usepackage{amssymb}
\usepackage{amsmath}
\usepackage{booktabs}
\usepackage{array}

\usepackage{rotating}

\usepackage[hidelinks]{hyperref}
\usepackage{xurl}
\usepackage{enumitem}
\setlist[enumerate,1]{leftmargin=1.25em,label=\alph*)}

\newcommand{\fitwidth}[1]{\resizebox{\ifdim\width>\linewidth\linewidth\else\width\fi}{!}{#1}}

\journal{}

\newif\ifanon
\newif\iftitlepage
\ifnum\pdfstrcmp{\jobname}{paper-anonymized}=0 \anontrue\fi
\ifnum\pdfstrcmp{\jobname}{paper-title-page}=0 \titlepagetrue\fi

\begin{document}

\begin{frontmatter}


\title{Human Values in a Single Sentence: Moral Presence, Hierarchies, and Transformer Ensembles on the Schwartz Continuum} 

\ifanon\else
\author[upv,uev]{Víctor Yeste\corref{cor}} 
\ead{vicyesmo@upv.es}
\author[upv,valgrai]{Paolo Rosso} 

\affiliation[upv]{organization={PRHLT Research Center, Universitat Politècnica de València},
            city={Valencia},
            postcode={46022},
            country={Spain}}

\affiliation[uev]{organization={School of Science, Engineering and Design, Universidad Europea de Valencia},
            city={Valencia},
            postcode={46010},
            country={Spain}}

\affiliation[valgrai]{organization={Valencian Graduate School and Research Network of Artificial Intelligence (ValgrAI)}}

\cortext[cor]{Corresponding author}
\fi

\begin{abstract}
We study neural multi-label classification under severe label imbalance through sentence-level detection of the 19 refined Schwartz human values in 74k English news and manifesto sentences (ValueEval'24 corpus). Each sentence carries a roughly balanced moral-presence label and a 19-way value annotation. First, moral presence is learnable from single sentences: a DeBERTa-base classifier reaches positive-class $F_1 \approx 0.73$ at the default threshold, which calibration does not improve. Second, comparing direct multi-label detectors with presence-gated hierarchies under an 8\,GB consumer-grade GPU budget, we find that gating does not improve over direct prediction, as gate recall becomes a bottleneck. Third, studying lightweight auxiliary signals and small ensembles, we isolate decision-threshold calibration as a decisive, often overlooked factor: a standard text-only baseline already matches the best official ValueEval'24 English run at the default threshold (macro-$F_1 = 0.282$ vs.\ $\approx 0.28$), and tuning the threshold on validation alone raises it to $0.315$, most of our overall gain. Lightweight features do not survive a paired per-seed test; a soft-voting ensemble reaches our best macro-$F_1 = 0.332$. Calibration is not architecture-specific: it reproduces on RoBERTa-base, where the gain is larger ($+0.043$). To our knowledge, this is the first systematic comparison of direct and presence-gated architectures, lightweight feature-augmented encoders, and instruction-tuned Large Language Models (LLMs) at sentence level; benchmarked 7--9B LLMs (zero-/few-shot and QLoRA) lag behind the supervised ensemble under the same budget. We provide empirical guidance for compute-efficient, value-aware NLP models.
\end{abstract}


\end{frontmatter}

\iftitlepage\else


\section{Introduction}
\label{sec:introduction}

Neural text classifiers for multi-label problems with many rare labels face an easily overlooked difficulty. The conventional decision rule, which assigns a label whenever its predicted probability exceeds $0.5$, systematically under-predicts infrequent classes, so gains attributed to new architectures or features can be confounded with how decisions are thresholded. Separating the two requires a task whose labels are numerous, severely imbalanced, and semantically close to one another. Sentence-level detection of human values is such a task, and it is the testbed for this study.

Human values are central to explaining and predicting attitudes, decisions, and behavior in individuals and groups \citep{Rokeach1973,HitlinPinkston2013}. Among the most influential frameworks, Schwartz's theory of basic human values models values as motivational goals arranged in a continuous circular structure, where adjacent values express compatible motivations and opposing values express conflicts \citep{Schwartz1992,Schwartz2012}. The refined model distinguishes 19 basic values (e.g., \emph{Self-direction: thought}, \emph{Security: societal}, \emph{Universalism: concern}), each linked to underlying needs and motivational emphases, and has been widely used in social psychology and political science to study value structure, value change, and cross-cultural invariance \citep{BardiSchwartz2003,Davidov2014}. Figure~\ref{fig:schwartz_continuum} illustrates this circular motivational continuum.

\begin{figure}[htbp]
    \centering
    \includegraphics[width=\linewidth]{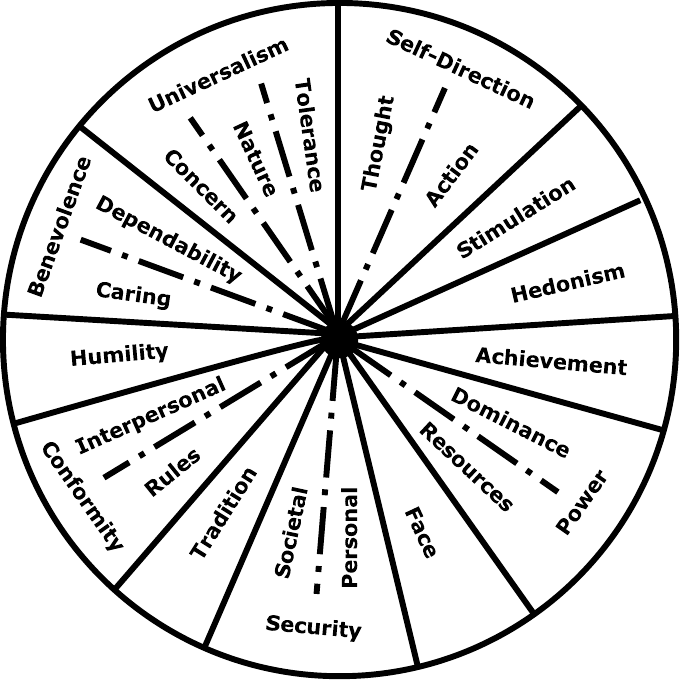}
    \caption{Circular motivational continuum of the 19 refined basic values in Schwartz's theory. Neighbouring values are motivationally compatible, whereas values on opposite sides of the circle tend to be in conflict. Adapted from \citet{Schwartz2012}.}
    \label{fig:schwartz_continuum}
\end{figure}

In parallel, work in Natural Language Processing (NLP) and computational social science has begun to infer moral or value-related content directly from text. Value-aware and morality-aware NLP has been applied to political communication \citep{Wickenkamp2025}, news and media analysis \citep{DIgnazi2025}, stance detection \citep{AlDayelMagdy2021}, and social discussions \citep{Preniqi2024}, and recent surveys systematise this emerging area and its benchmarking challenges \citep{Rink2025}. Much of this research has drawn on Moral Foundations Theory (MFT) \citep{HaidtGraham2007,Graham2013}, which proposes a small set of intuitive moral foundations and associated resources such as the Moral Foundations Questionnaire \citep{Graham2011} and the Moral Foundations Dictionary and its extensions \citep{Hopp2021,Hoover2020}. These resources support supervised and lexicon-based models that detect whether a text invokes, for example, harm, fairness, or loyalty \citep{Rezapour2021,Trager2022}. 

Using Schwartz's value theory in NLP is more recent and less explored than MFT, but it is particularly suitable for political and ideological language. Schwartz values have a long measurement tradition in survey research, for example with the Portrait Values Questionnaire (PVQ) \citep{Schwartz2001}, strong evidence of cross-cultural robustness \citep{Davidov2008,Schwartz2016}, and well-documented links to political preferences and issue positions \citep{Caprara2006,Ros1999}. Recent projects have begun to operationalise the refined 19-value continuum for text, most prominently in the ValuesML collection and the ValueEval shared tasks \citep{ValueEval24Zenodo,Kiesel2024,Kiesel2023}. These initiatives provide sentence- or segment-level annotations of value presence and thus make it possible to study fine-grained value detection with modern NLP models.

Detecting values at the sentence level, however, remains difficult. First, value cues in naturalistic text are often subtle or implicit, especially in technocratic or institutional language, and may only become clear when surrounding context is considered \citep{Lemke1990}. Second, the distribution of values is highly skewed: values such as \emph{Security: societal} or \emph{Conformity: rules} are common in political and news texts, whereas others such as \emph{Humility} or \emph{Hedonism} are rare \citep{ValueEval24Zenodo}. This leads to severe class imbalance and makes macro-averaged metrics particularly demanding. Third, models must distinguish between neighbouring values on the continuum (e.g., \emph{Benevolence} vs.\ \emph{Universalism}) and between different facets of similar themes (e.g., \emph{Power: dominance} vs.\ \emph{Power: resources}). This requires fine-grained semantic distinctions that go beyond generic sentiment or topic information \citep{Schwartz2012}.

Modern transformer-based encoders and instruction-tuned Large Language Models (LLMs) provide strong tools for such tasks. Contextual encoders such as BERT \citep{Devlin2019}, RoBERTa \citep{Liu2019}, and DeBERTa \citep{He2021} have been successfully applied to moral foundation detection, political framing, and stance classification \citep{Timoneda2025, Zhang2025}. Instruction-tuned LLMs can perform zero- or few-shot classification from natural-language label descriptions \citep{Ouyang2022,Wei2022,Gilardi2023} and can also generate text under moral or value-related constraints, for example in humour generation conditioned on moral judgments \citep{Yamane2021}. At the same time, recent evaluations suggest that medium-sized encoder models, when carefully fine-tuned and combined with simple ensembles, can match or surpass much larger LLMs on structured multi-label problems under realistic compute constraints \citep{Liang2023}.

\subsection{Research objectives}
\label{sec:objectives}

In this paper, we study sentence-level identification of the 19 values in the refined Schwartz continuum as a concrete formulation of human value detection. We focus on the English, machine-translated release of the dataset used in the ValueEval'24 shared task \citep{ValueEval24Zenodo,Kiesel2024}, where each sentence from news articles or political manifestos is annotated for the presence of each value and for whether the value is portrayed as attained or constrained. Following prior work on value detection \citep{Kiesel2023,Kiesel2024}, we collapse the \emph{stance} information (attained vs.\ constrained) into a single binary label per value and \emph{keep} the full 19-dimensional label space. In addition, we define a derived \emph{moral presence} variable indicating whether any value is annotated as present in a sentence. This yields two related prediction problems that we study \emph{in parallel}, rather than reducing one to the other: (i) binary detection of moral presence and (ii) multi-label detection of the 19 values.

Our overarching goal is to understand how well current transformer-based and instruction-tuned language models can detect refined Schwartz values at the sentence level under strong class imbalance, targeting a deployment setting where only a single consumer-grade GPU is available. We structure this goal into four research questions (RQ1--RQ4):

\begin{enumerate}[label=RQ\arabic*.,leftmargin=2em]
  \item Can we reliably detect the presence of moral content in single sentences?
  \item Does a moral-presence gate (hierarchical) help over a direct multi-label value detector on the 19-way Schwartz value detection task, under matched compute?
  \item Which lightweight auxiliary signals help under a fixed single-GPU compute budget?
\item How do supervised DeBERTa models compare to instruction-tuned open LLMs (7--9B parameters) and their ensembles?
\end{enumerate}

Throughout, the 19-way multi-label task is our main target: the presence variable $z_s$ is introduced as an auxiliary signal with its own evaluation (RQ1) and as a potential gate in hierarchical architectures (RQ2), but we always report and discuss results on the full 19-value prediction problem.

We make three main contributions. First, we operationalise a sentence-level \emph{moral presence} task on the ValueEval'24 data and show that it is learnable with positive-class $F_1$ around 0.73 using DeBERTa-based classifiers, despite sparse and often implicit cues. Second, we perform a controlled comparison between direct multi-label value detectors and hierarchical pipelines in which a presence gate filters sentences before value prediction. Under a realistic 8\,GB GPU budget, we find that presence gating does not clearly outperform direct prediction, suggesting that gate recall can become a bottleneck for downstream values. Third, we systematically explore lightweight auxiliary signals---short-range context, psycholinguistic and moral lexica, and topic features---and small ensembles of supervised encoders and instruction-tuned LLMs, and we separate their effect from that of decision-threshold calibration. We show that, under the severe class imbalance of this task, calibrating the decision threshold of a standard baseline accounts for essentially all of the improvement: the auxiliary signals we tested add at most $+0.005$ macro-$F_1$ on average, which a paired per-seed test cannot separate from seed variance, while our best soft-voting ensemble of DeBERTa-based models still outperforms individual models and medium-scale LLM baselines.

By grounding value detection in the refined Schwartz continuum and by comparing direct, hierarchical, and ensemble architectures under explicit resource constraints, this work provides empirical evidence and practical guidance for building value-aware NLP models. Although all experiments are conducted on the English, machine-translated portion of the ValueEval'24 corpus, the comparisons we draw---between presence-gated and direct architectures, between lightweight features and bare encoders, and between single models and small ensembles---are methodological in nature, and we expect them to be informative for other languages and domains with similarly fine-grained, imbalanced value taxonomies; we treat this broader transferability as a hypothesis for future work rather than an established result, since all our experiments rest on a single corpus, a single encoder family, and one LLM scale band. Beyond this specific dataset, our findings highlight the often-underestimated role of decision-threshold calibration---which by itself recovers most of the improvement over prior baselines under class imbalance---together with lightweight features and ensembling for fine-grained, imbalanced moral classification, and suggest that, within the modest single-GPU budget considered here and at the 7--9B parameter scale, moderately sized supervised encoders remain a strong and compute-efficient baseline relative to instruction-tuned LLMs for structured human value detection.

To keep the main text focused, extended tables, per-value breakdowns, ablation studies, and additional implementation details that were previously in the appendices are now collected in the Supplementary Material accompanying this article.

This article adopts the best-performing English run on the ValueEval'24 corpus \citep{Yeste2024}---a cascade system submitted to the shared task---as its primary supervised baseline, and substantially broadens and generalises the analysis beyond that single system. That shared-task system introduced a cascade model for English value detection on the same underlying corpus and reported competitive results in the official competition setting. In our experiments we reimplement the direct DeBERTa-based component of \citet{Yeste2024} on the official ValueEval'24 English splits and use it as the starting point for all supervised baselines (Section~\ref{sec:baselines}). In contrast, the present work (i) formalises and evaluates a sentence-level \emph{moral presence} task alongside the 19-label value prediction problem, (ii) systematically compares direct and presence-gated architectures, (iii) studies the impact of lightweight auxiliary signals such as short-range context, psycholinguistic and moral lexica, and topic features, (iv) benchmarks instruction-tuned LLMs (zero-/few-shot and quantized LoRA/QLoRA) and their ensembles against supervised encoders, and (v) conducts a more extensive statistical and error analysis. All experiments are re-run from scratch with a unified protocol and explicit significance testing, and the shared-task configuration of \citet{Yeste2024} is treated as the direct architecture baseline within this broader evaluation. Table~\ref{tab:prior-vs-this} summarises these differences; other ValueEval'24 shared-task systems may have explored some of these choices informally, so our contribution is the \emph{systematic}, controlled comparison under a fixed corpus, protocol, and hardware budget rather than priority over any individual design.

\begin{table*}[t]
\centering
\caption{Relationship between the prior conference system \citep{Yeste2024} and the present work, on the same English ValueEval'24 corpus and official splits.}
\small
\begin{tabular}{p{.18\linewidth} p{.35\linewidth} p{.35\linewidth}}
\toprule
\textbf{Aspect} & \textbf{\citet{Yeste2024}} & \textbf{This work} \\
\midrule
Tasks & 19-value detection & Binary moral presence \emph{and} 19-value detection \\
Architecture & DeBERTa-based cascade with competition-specific post-processing & Direct vs.\ presence-gated DeBERTa under one unified protocol \\
Auxiliary signals & None & Short-range context, LIWC-22, moral/affective lexica, topics (ablated) \\
LLM baselines & None & 7--9B instruction-tuned LLMs (zero-/few-shot, QLoRA) and ensembles \\
Compute budget & Not explicitly constrained & Explicit single 8\,GB GPU budget \\
Significance testing & Official competition ranking & Paired bootstrap, McNemar, Benjamini--Hochberg \\
Best English macro-$F_1$ & $0.28$ (best official run) & $0.332$ (DeBERTa ensemble); $0.315$ from threshold calibration of the baseline alone \\
\bottomrule
\end{tabular}
\label{tab:prior-vs-this}
\end{table*}

The remainder of this paper is organized as follows. Section~\ref{sec:related-work} reviews related work on human values and moral or value-related content detection in NLP and computational social science. Section~\ref{sec:task-data} introduces the ValueEval'24 corpus, formulates the presence and multi-label value prediction tasks, and summarises key descriptive statistics. Section~\ref{sec:methods} details our modelling approaches, including direct and hierarchical DeBERTa-based classifiers, lightweight auxiliary signals, instruction-tuned LLM baselines, and ensemble architectures. Section~\ref{sec:experimental-setup} specifies the experimental protocol, training and thresholding procedures, evaluation metrics, and hardware constraints. Section~\ref{sec:results} presents and discusses the empirical results for our four research questions, including a summary of findings, error analysis, limitations, and ethical considerations. Finally, Section~\ref{sec:conclusions-future-work} offers concluding remarks and outlines directions for future work.

\section{Related work}
\label{sec:related-work}

Research on computational modelling of human values connects three strands: (i) psychological theories and lexical resources for values and morality; (ii) sentence-level models for value or moral-content detection; and (iii) recent work with instruction-tuned Large Language Models (LLMs) and low-resource strategies. We review each strand with a focus on settings closest to our sentence-level formulation.

\subsection{Human values, moral frameworks, and lexical resources}
\label{sec:human-values-moral-frameworks}

Early work in psychology views values as enduring beliefs about desirable modes of conduct or end-states that guide attitudes and behaviour \citep{Rokeach1973}. Schwartz's theory of basic human values refines this view into a structured set of motivational goals organized in a circumplex continuum \citep{Schwartz1992,Schwartz2012}. The refined model distinguishes 19 basic values such as \emph{Self-direction: thought}, \emph{Security: societal}, and \emph{Universalism: concern}, and has been validated across cultures with instruments like the Portrait Values Questionnaire (PVQ) \citep{Schwartz2001}. This framework is widely used in political science to explain ideological orientations and policy preferences \citep{Caprara2006,Ros1999}.

In computational linguistics, much early work on moral or value-related language builds on Moral Foundations Theory (MFT) \citep{HaidtGraham2007,Graham2013}. MFT posits a small set of intuitive moral foundations (e.g., care/harm, fairness/cheating, authority/subversion) and has inspired resources such as the Moral Foundations Dictionary and its updates \citep{Hoover2020,Hopp2021}. These lexica map words to foundation categories (virtue/vice) and support lexicon-based and supervised analyses of moral rhetoric in news, social media, and political communication \citep{Rezapour2021,Trager2022,AlDayelMagdy2021}. Extended resources such as the extended Moral Foundations Dictionary (eMFD) \citep{Hopp2021}, the Moral Foundations Twitter Corpus \citep{Hoover2020}, and domain-specific lists (e.g., for radicalisation or hate speech) broaden coverage but typically operate at the level of broad moral themes rather than fine-grained value distinctions. For a recent overview of NLP work on morality in text, see \citet{Reinig2024}.

Closer to Schwartz values, recent projects have begun to operationalise the refined 19-value continuum for text. ValuesML \citep{ValueEval24Zenodo} introduces a multilingual corpus of news and political texts annotated for Schwartz values and underlies the ValueEval shared tasks (ValueEval'23 at SemEval~2023 and ValueEval'24 at the Touch\'e lab, CLEF~2024) \citep{Kiesel2023,Kiesel2024}. These tasks provide sentence- or segment-level annotations for the 19 values (and higher-order groups), enabling direct multi-label classification on the refined continuum. Beyond ValueEval, work in computational social science has used Schwartz-inspired value lexica and regression models to estimate value profiles from social media posts or political speeches, often at the user or document level \citep{Jahanbakhsh2025}. Further efforts explore value elicitation and modelling in applied settings, such as interactive value promotion schemes \citep{GarciaRodriguez2025}, showing the potential of Schwartz-based representations for decision support and policy analysis.

Very recent work generalises value and morality classification beyond a single theoretical framework. \citet{Chen2025} introduce MoVa, a benchmark suite with 16 labeled datasets and four value frameworks, and show that carefully designed multi-label prompting strategies can transfer across domains and label taxonomies. \citet{Borenstein2025} and \citet{Starovolsky-Shitrit2025} study human values in online communities and short-video platforms, combining large-scale text (and multimodal) analysis with psychologically grounded value taxonomies. These efforts complement ValuesML and ValueEval by situating value detection within broader computational social-science workflows and by illustrating demand for value-aware NLP tools in real media and platform settings. Compared to MFT-based resources, Schwartz-based datasets use a richer, continuous value space with more labels, many of them rare, which makes sentence-level value detection particularly challenging.

\subsection{Sentence-level value and moral content detection}
\label{sec:sentence-level}

Within NLP, moral and value-related content has been modelled at different granularities and with different label taxonomies. MFT-based studies often treat moral foundation detection as a sentence- or tweet-level multi-label task. \citet{Rezapour2021}, for example, use BERT-based classifiers to detect moral foundations in political tweets and show that contextual embeddings outperform lexicon-only baselines but still struggle with subtle or implicit moral language. \citet{Trager2022} analyse moral rhetoric in media by combining foundation detection with topics and stance, and find systematic interactions between moral framing, party, and issue domain.

The ValueEval shared tasks \citep{Kiesel2023,Kiesel2024} shift the focus from moral foundations to Schwartz values. ValueEval'23 \citep{Kiesel2023}, run at SemEval~2023, targeted the identification of human values behind \emph{arguments} (premise--conclusion pairs with stance), using the Touch\'e23-ValueEval dataset \citep{Mirzakhmedova2024}. ValueEval'24 \citep{Kiesel2024}, run at the Touch\'e lab at CLEF~2024, moved to \emph{news-article and political-manifesto sentences} annotated with the 19 refined values (and aggregate groups) in several languages, with explicit stance labels (attained vs.\ constrained); this is the edition and corpus we use \citep{ValueEval24Zenodo}. Participants typically used transformer-based encoders (e.g., mBERT, XLM-R, DeBERTa) in multi-label setups, sometimes augmented with lexical or topic features. In the English-only setting, reported macro-$F_1$ scores remain modest, with the best team reaching $\approx 0.28$ \citep{Yeste2024}, reflecting strong label imbalance and the difficulty of distinguishing closely related values. This limits the reliability of fully automated tools for fine-grained analysis of political rhetoric or value appeals in news. \citet{Rink2025} provide a broader benchmarking perspective and reach similar conclusions about the difficulty of human-value detection benchmarks.

Beyond ValueEval, recent work proposes architectures for human value identification tailored to large-scale applications. EAVIT \citep{Zhu2025}, for example, uses LLMs as value scorers within an efficient pipeline and reports strong performance on multiple value benchmarks. Together with MoVa \citep{Chen2025} and the survey by \citet{Rink2025}, these studies underline that automated value detection is increasingly treated as a core text-mining task, while also confirming that fine-grained, sentence-level labels with strong class imbalance remain challenging.

Related work on political framing \citep{Card2015}, newspaper editorials \citep{Kiesel2015}, propaganda detection \citep{Yu2021}, and ideological rhetoric \citep{Pan2024} addresses sentence- or clause-level prediction of higher-level categories such as frames, issues, or stance. These tasks also rely on transformer encoders and may incorporate lexical cues (e.g., LIWC \citep{TausczikPennebaker2010}) or domain-specific dictionaries, but target coarser label spaces than the 19-value continuum. Our formulation follows ValueEval in treating each sentence as an independent unit and asking for the presence of any of the 19 refined values, which amplifies the effect of subtle cues and class imbalance. At the same time, this fine-grained, sentence-level perspective is what is needed to study how political actors and media outlets foreground different values across clauses and sentences, and to support nuanced, value-aware analyses of political discourse.

\subsection{Pipelines, hierarchies, and contextual cues}
\label{sec:pipelines-hierarchies}

A natural question in structured prediction is whether intermediate tasks or hierarchical pipelines help over direct multi-label prediction. In moral and value-related NLP, this often means first deciding whether a text contains \emph{any} moral content and then predicting specific categories only for morally positive texts. In MFT-based detection, some authors implicitly adopt this structure by collapsing all foundations into a binary moral-vs.-non-moral label and then fine-tuning foundation-specific models on the subset of moral texts \citep{Trager2022}. However, systematic comparisons between such gated pipelines and direct multi-label models are rare, and reported gains are mixed once models are calibrated and trained under comparable resource constraints.

Hierarchical architectures are common in related domains such as document classification and legal text analysis, where sentence-level modules feed into document-level decisions \citep{Yang2016,Chen2023}. Multi-label text classifiers with many labels also model label dependencies directly, for example by jointly learning label embeddings and label correlations \citep{Liu2021}. In these settings, context aggregation (for example via attention over sentences) is central. For sentence-level detection, context is usually injected more simply, for example by concatenating neighbouring sentences or paragraphs or by using sliding windows over a document \citep{Kuparinen2023}. These context windows add topical or rhetorical information (e.g., mentions of policies, groups, or harms) that can disambiguate otherwise neutral sentences, but they also increase sequence length and VRAM usage, which matters under tight hardware budgets.

An orthogonal strategy is to augment encoders with auxiliary features such as psycholinguistic lexica, emotion or morality scores, or topic distributions. LIWC categories capture psychological dimensions related to affect, social processes, and cognitive style \citep{Pennebaker2015} and have been linked to personality, political orientation, and moral rhetoric \citep{PreotiucPietro2017}. In moral foundation detection, combining BERT or RoBERTa representations with lexicon-derived features (e.g., MFD counts, eMFD scores, or domain-specific dictionaries) yields small but consistent gains, especially for rare foundations \citep{Rezapour2021}. Topic-based features, drawn from classical Latent Dirichlet Allocation (LDA) or Non-negative Matrix Factorization (NMF) \citep{Jelodar2019,LeeSeung1999}, or from neural topic models such as BERTopic \citep{Grootendorst2022}, have likewise been used as lightweight contextual cues in political and news classification \citep{Kiesel2015}. Our work follows this tradition: we explore short-range context, psycholinguistic and moral lexica, and topic features as compute-frugal signals attached to a DeBERTa encoder and evaluate them under an explicit 8\,GB GPU budget and in both direct and presence-gated architectures.

\subsection{LLMs, prompting, and low-resource strategies}
\label{sec:llms-prompting}

Instruction-tuned LLMs offer an alternative route to moral and value-related classification. Early studies used GPT-3 style models in zero-shot or few-shot setups to detect moral foundations or political stances from natural-language label descriptions \citep{Schramowski2022,Gilardi2023}. Beyond classification, LLMs have been prompted to generate text that conforms to moral or value constraints, for example in joke generation conditioned on moral judgments \citep{Yamane2021}. Recent evaluations on human-value benchmarks report that LLMs can capture broad moral distinctions and reach competitive performance, but often require large models and still show more variance across value categories than specialised encoders \citep{Sun2024}. Results suggest that LLMs internalise broad moral and political distinctions and can perform reasonably well without task-specific fine-tuning, but they may underperform specialised encoders on fine-grained or heavily imbalanced labels.

A parallel line of work studies the values and value alignment of LLMs themselves, often directly using Schwartz or related frameworks. ValueFULCRA \citep{Yao2024}, for example, maps LLM outputs into a multidimensional basic-value space and proposes a value-alignment paradigm based on Schwartz values. Other studies develop psychometric methods for measuring human and AI values from free-text outputs \citep{Ye2025}, analyse the consistency of LLM value profiles \citep{Rozen2025}, or compare human and model values across cultures and scenarios \citep{Shen2025}. These works treat LLMs as \emph{objects} of value measurement, whereas our focus is on using both medium-sized encoders and 7--9B instruction-tuned LLMs as \emph{tools} for detecting refined Schwartz values in human-authored sentences under explicit resource constraints.

Instruction-tuned open models such as Llama, Gemma, or Qwen, combined with parameter-efficient fine-tuning (e.g., LoRA/QLoRA), provide a middle ground between full supervised training and pure prompting \citep{Dettmers2024,Hu2022}. For multi-label sentence classification, QLoRA adapters can be trained on commodity GPUs while keeping the base model frozen and are increasingly used for domain adaptation in legal, biomedical, programming, and social-science tasks \citep{Venkatesh2025}. However, recent benchmarks indicate that medium-sized encoder models (BERT, RoBERTa, DeBERTa), when carefully fine-tuned and combined with simple ensembles, can match or surpass considerably larger LLMs on structured multi-label problems, especially when compute and data are limited \citep{Liang2023}.

Ensembling is a long-standing technique in NLP and machine learning \citep{Dietterich2000} and is particularly helpful for imbalanced multi-label classification \citep{Tsoumakas2011}. For transformer-based sentence classifiers, simple probability- or majority-vote ensembles over independently trained models can improve robustness and macro-averaged metrics \citep{Tahir2012}. In moral and value detection, studies have reported that combining encoders with lexicon-based models or with different random seeds yields more stable performance on rare categories \citep{Rezapour2021,Hoover2020}. Label imbalance can also be handled outside the model architecture. \citet{Charte2015} propose measures of imbalance for multi-label datasets together with resampling algorithms that reduce it, and \citet{Collell2018} show that a bagging ensemble combined with threshold moving, which sets the decision threshold after training for the metric of interest, is competitive with rebalancing approaches. Our work contributes to this line by comparing (i) supervised DeBERTa models with lightweight features, (ii) instruction-tuned LLMs used via prompting and QLoRA, and (iii) small ensembles that mix models across these families, all under matched data and an explicit 8\,GB single-GPU constraint.

In summary, existing research shows that sentence-level moral and value detection is feasible with both encoder-based models and LLMs, but several questions remain open. In this paper we focus on three of them: (i) the empirical value of presence gating versus direct prediction on the refined Schwartz continuum; (ii) the extent to which lightweight context, lexica, and topics help under tight VRAM budgets; and (iii) how carefully tuned supervised encoders compare to instruction-tuned LLMs and their ensembles on a fine-grained, imbalanced value taxonomy.

\section{Task and data}
\label{sec:task-data}

\subsection{Task definition}
\label{sec:task-definition}

We adopt the refined Schwartz value continuum with 19 basic values \citep{Schwartz2012}. For each sentence $s$ and value $v \in \mathcal{V}$, the gold annotation provides two stance indicators, attainment and constraint, denoted
\[
\mathrm{attained}_{s,v},\ \mathrm{constrained}_{s,v} \in \{0,1\}.
\]
We collapse them into a single binary label
\[
y_{s,v} = \mathbb{I}\!\left[\,\mathrm{attained}_{s,v} + \mathrm{constrained}_{s,v} > 0\,\right],
\]
and define the \emph{moral presence} variable
\[
z_s = \mathbb{I}\!\left[\,\exists\, v \in \mathcal{V}: y_{s,v} = 1\,\right].
\]

We therefore study two sentence-level prediction problems:
(i) binary detection of moral presence ($z_s$), and
(ii) multi-label detection of the 19 values ($\{y_{s,v}\}_{v\in\mathcal{V}}$).
Unless stated otherwise, value detectors are evaluated with macro-averaged $F_1$ over the positive class across the 19 labels; per-label metrics are reported in the Supplementary Material.

We treat moral presence as a separate prediction problem for three reasons. First, presence is a useful signal in its own right: a reliable filter of value-rich sentences is helpful for downstream qualitative analysis and for prioritising instances in annotation workflows. Second, many practical systems naturally adopt a hierarchical structure in which a presence detector gates more expensive, fine-grained value classifiers; our experiments explicitly test whether such a gate is beneficial under realistic compute constraints. Third, from a modelling perspective it is informative to ask to what extent the mere \emph{existence} of value-related content can be detected from a single sentence, independently of which specific values are active. Importantly, throughout the paper we always evaluate presence and the 19 value labels separately and never replace the multi-label value task by the presence task.

Figure~\ref{fig:label_space} summarises this formulation. Each instance consists of a sentence $s$, its 19-dimensional value vector $\mathbf{y}(s)$, and the derived moral presence label $z_s$.

\begin{figure}[htbp]
    \centering
    \includegraphics[width=\linewidth]{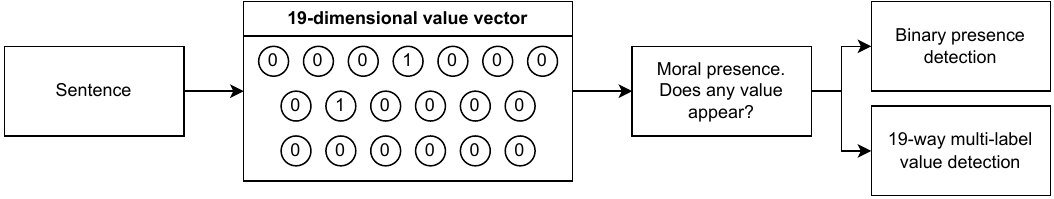}
    \caption{Sentence-level label space and prediction tasks. Each sentence $s$ is associated with a 19-dimensional value vector $\mathbf{y}(s)$, with one component for each refined Schwartz value listed in Section~\ref{sec:task-definition}. For readability, the figure shows the vector as a single block rather than labelling each individual value.}
    \label{fig:label_space}
\end{figure}

\subsection{Dataset}
\label{sec:dataset}

We use the English, machine-translated release of the ValueEval'24 dataset prepared for the Touch\'e lab at CLEF 2024, derived from the ValuesML project \citep{ValueEval24Zenodo,Kiesel2024}. The underlying ValuesML corpus consists of around 3{,}000 human-annotated news articles and political manifestos (roughly 400--800 words) on contemporary policy topics. As part of the ValueEval shared tasks on human value detection \citep{Kiesel2023,Kiesel2024}, these documents were segmented into sentences and annotated with the 19 refined Schwartz values.

The organizers distribute sentence-level labels in nine languages and a machine-translated English version. In this paper we work exclusively with the English, machine-translated sentences and the official train/validation/test split provided for Subtask~1 (sentence-level value detection) of ValueEval'24. The English subset contains 44{,}758 training, 14{,}904 validation, and 14{,}569 test sentences (74{,}231 in total), of which roughly half express at least one value (Section~\ref{sec:stats}).

These sentences derive from 2{,}648 source documents whose original genre and language are preserved in the document identifiers, allowing us to characterise the release we use. By genre the corpus is dominated by news: 2{,}354 documents (88.9\%) are news articles and 294 (11.1\%) political manifestos, accounting for 90.1\% and 9.9\% of sentences respectively. Although the release is in English, only about 14\% of its sentences originate from English texts; the remainder is machine-translated from eight other source languages. By sentence count the source-language composition is Turkish (15.0\%), Dutch (14.8\%), English (13.9\%), German (12.4\%), Greek (9.9\%), Hebrew (9.9\%), Bulgarian (9.3\%), Italian (8.6\%), and French (6.3\%). This predominance of machine-translated content is relevant to the lexicon-based features and is discussed further in Section~\ref{sec:limitations-ethics}.

Annotation follows the official ValuesML guidelines for values in news and political manifestos \citep{ValuesGuidelines2024}. Annotators work primarily at the sentence level: for each sentence they highlight the minimal span that expresses a value, assign one or more of the 19 refined Schwartz values, and label whether the value is (partially) attained, (partially) constrained, or not coded.

For example, the guidelines annotate the sentence
\begin{quote}
We need to do more to protect the environment.
\end{quote}
with the value \emph{Universalism: nature} and an attainment label \emph{(partially) attained}, because it is a call to act in order to safeguard the natural environment. Likewise, the sentence
\begin{quote}
We should be happy and satisfied with what we have.
\end{quote}
is annotated as expressing \emph{Humility} with \emph{(partially) attained}, since it urges contentment with one’s current situation \citep[Section~3.3]{ValuesGuidelines2024}.

In our experiments (Section~\ref{sec:task-definition}) we collapse the attainment information into a single binary label per value and derive the sentence-level \texttt{presence} variable $z_s$ indicating whether at least one value is active.

\paragraph{Licensing and access}
The dataset is distributed under a restricted Data Usage Agreement: it may be used for scientific research on human value detection, but redistribution (in part or in full) is prohibited. Access is via Zenodo \citep{ValueEval24Zenodo}. To comply with this license, we release only our \emph{code, configurations, tuned thresholds, and per-model predictions}, not the texts themselves (Section~\ref{sec:data-availability}).

\paragraph{Data format and alignment}
Sentences are keyed by \texttt{Text-ID} and \texttt{Sentence-\allowbreak ID} in \texttt{sentences.tsv}. Labels reside in \texttt{labels-cat.tsv} with 19 value columns (binary), plus a \texttt{presence} column consistent with $z_s$. Evaluation always merges predictions with gold labels on (\texttt{Text-ID}, \texttt{Sentence-ID}) to guarantee one-to-one alignment.

\subsection{Descriptive statistics}
\label{sec:stats}

Table~\ref{tab:corpus-stats} reports sentence counts and the share of sentences with any value (\%~presence) per split. Overall, around half of the sentences express at least one value.

\begin{table*}[t]
\centering
\caption{Corpus statistics by split (English, machine-translated).}
\begin{tabular}{lrr}
\toprule
\textbf{Split} & \textbf{\# Sentences} & \textbf{\% Presence} \\
\midrule
Train      & 44{,}758 & 51.53 \\
Validation & 14{,}904 & 50.99 \\
Test       & 14{,}569 & 50.81 \\
\bottomrule
\end{tabular}
\label{tab:corpus-stats}
\end{table*}

The distribution across the 19 values is highly imbalanced. Frequent values such as \emph{Security: societal} occur in almost 8--9\% of sentences across splits, whereas others, such as \emph{Self-direction: thought}, \emph{Universalism: tolerance}, or \emph{Humility}, appear in less than 3\%. This imbalance motivates the use of macro-averaged metrics, threshold calibration, and careful significance testing discussed in later sections. Full per-value prevalences per split are provided in \ref{app:prevalence}.

\section{Methods}
\label{sec:methods}

\subsection{Notation and decision rules}
\label{sec:notation-thresholds}

Let $\mathcal{V}$ be the set of 19 refined Schwartz values and $s$ a sentence. A value detector outputs a probability $\hat{p}_{s,v} \in [0,1]$ for each $v \in \mathcal{V}$. We evaluate models using macro-averaged $F_1$ over the positive class across all 19 values.

To obtain binary predictions, we apply value-specific thresholds
\[
\hat{y}_{s,v} = \mathbb{I}[\hat{p}_{s,v} \ge \tau_v].
\]
We consider two thresholding schemes:

\begin{itemize}[leftmargin=1.5em]
  \item \textbf{Fixed global threshold.} A single $\tau \equiv 0.5$ applied to all 19 values.
  \item \textbf{Tuned global threshold.} A single $\tau^\star \in [0,1]$, swept on the validation set to maximise macro-$F_1$, then frozen and applied to all 19 values on the test set. This is the scheme used for every tuned 19-value result reported in Sections~\ref{sec:results-hierarchy}--\ref{sec:results-llms} (the single $\tau^\star$ is reported in each table).
\end{itemize}

We also use the \emph{label-wise} scheme, in which each label's threshold $\tau_v$ is chosen on validation to maximise positive-class recall subject to a minimum precision of $0.40$ (searching $\tau_v \in [0.10, 0.90)$). This label-wise scheme is what we apply to the binary presence gate (Section~\ref{sec:results-presence}) and to the instruction-tuned LLM baselines (Section~\ref{sec:results-llms}); the DeBERTa value detectors and their ensembles instead use a single global threshold for the reported 19-value results (Tables~\ref{tab:direct-vs-hier}--\ref{tab:transformers-vs-llms}), which keeps the decision rule simple and avoids tuning 19 separate thresholds. Because the LLM outputs are close to binary, label-wise tuning has little effect on them in practice.

For the binary presence task, models output a probability $\hat{p}_s$ for \texttt{presence}$=1$. We again map to $\hat{z}_s \in \{0,1\}$ using a scalar threshold $t$ (fixed at $0.5$ or tuned on validation with the label-wise scheme above; see Section~\ref{sec:thresholding}).

\subsection{Model overview}
\label{sec:model-overview}

Figure~\ref{fig:model_overview} summarises the model families we evaluate:

\begin{enumerate}[label=\alph*),leftmargin=2em]
  \item \textbf{Direct multi-label DeBERTa-base value detectors} predict the 19 values from a single sentence, optionally enriched with lightweight auxiliary signals: prior-sentence context and labels, psycholinguistic and moral lexica, and topic features.
  \item \textbf{Presence-gated pipelines} first predict a binary moral presence label $z_s$ and only apply the value detector to sentences predicted as moral. A hard gate at threshold $\tau_{\text{gate}}$ zeroes out all value probabilities for sentences below the gate.
  \item \textbf{Instruction-tuned LLMs} (zero-/few-shot prompting and QLoRA) take a sentence and a natural-language description of the 19 values and return a set of active values, which we map to a binary vector.
  \item \textbf{Ensembles} combine predictions from multiple supervised DeBERTa models and, in some variants, LLM-based classifiers via soft or hard voting.
\end{enumerate}

\begin{figure}[htbp]
    \centering
    \includegraphics[width=0.75\linewidth]{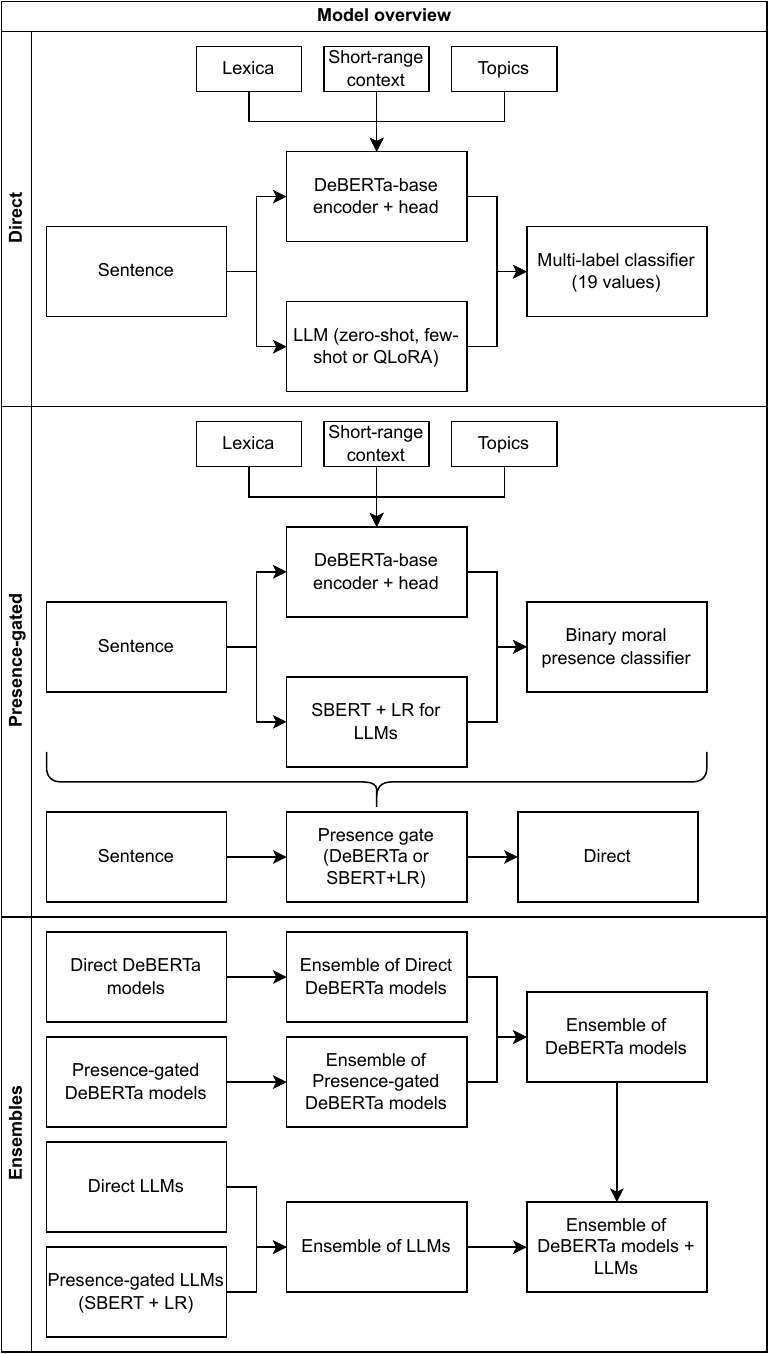}
    \caption{Overview of the model families considered in this work.}
    \label{fig:model_overview}
\end{figure}

All models are trained on the training split, tuned on the validation split, and finally evaluated on the held-out test split. Thresholds and hyperparameters are always selected on validation. Figure~\ref{fig:selection_evaluation} shows this selection and evaluation process.

To avoid ambiguity, we explicitly separate the training objectives for the two tasks introduced in Section~\ref{sec:task-definition}. Models for \emph{moral presence} are trained as binary classifiers on the scalar label $z_s$ only. Models for the 19 values are trained as multi-label classifiers on the vector $\{y_{s,v}\}_{v\in\mathcal{V}}$ only. We never train a single model that predicts presence instead of the 19 values, and all results on the value task are based on models that directly optimise the 19-dimensional label vector.

This decomposition into model families is deliberate: by varying only a small number of architectural choices (direct vs.\ presence-gated, encoder vs.\ LLM, single model vs.\ compact ensemble) under a unified training and hardware setup, we turn the comparison itself into a methodological contribution about how to build value-aware NLP systems in practice.

\begin{figure}[htbp]
    \centering
    \includegraphics[width=0.75\linewidth]{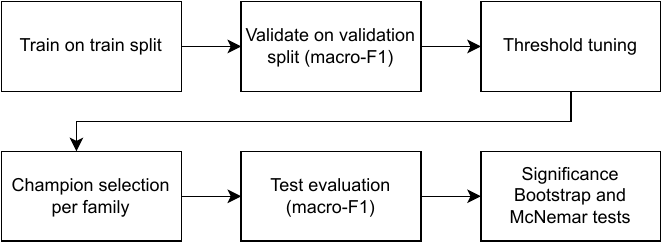}
    \caption{Model selection and evaluation: threshold tuning, champion selection, and test evaluation.}
    \label{fig:selection_evaluation}
\end{figure}

\subsection{Baselines and comparison protocol}
\label{sec:baselines}

Our comparisons are anchored to strong baselines trained under the same data and compute conditions.

\paragraph{Supervised DeBERTa baseline}
We use a DeBERTa-base direct multi-label classifier on the English, machine-translated ValueEval'24 sentences with the official train/validation/test split (Section~\ref{sec:dataset}). Architecturally, this corresponds to the direct DeBERTa component of the cascade model in \citet{Yeste2024}, but without the additional rule-based or competition-specific post-processing. We reimplement and retrain this configuration so that all supervised baselines and proposed variants share the same codebase, optimisation settings, and thresholding protocol.

\paragraph{Instruction-tuned LLM baselines}
As generative baselines, we consider four 7--9B instruction-tuned open models that fit on a single 8\,GB GPU: Gemma~2~9B \citep{Gemma2}, Llama~3.1~8B \citep{Llama3}, Ministral~8B \citep{Ministral8B}, and Qwen~2.5~7B \citep{Qwen25}. These models have shown strong performance on a range of classification tasks when used with prompting or parameter-efficient fine-tuning. In all cases we use exactly the same ValueEval'24 English splits and label definitions as for DeBERTa.

Unless explicitly stated otherwise (e.g., when quoting the official ValueEval leaderboard score of \citet{Yeste2024} for context), all results reported in Section~\ref{sec:results} come from models that we run on the ValueEval'24 English splits under a single 8\,GB GPU constraint.

\subsection{Direct multi-label baseline (DeBERTa-base)}
\label{sec:direct}

Our main supervised encoder is \texttt{microsoft/deberta-base} \citep{He2021}, fine-tuned with a linear multi-label head and \textsc{BCEWithLogits} loss. This choice follows a five-way comparison of pretrained encoders reported on the same corpus for the shared task \citep{Yeste2024}, in which DeBERTa outperformed BERT, RoBERTa, ELECTRA and XLNet; we revisit the choice in Section~\ref{sec:results-lightweight}, where we repeat the central calibration experiment with a second encoder family. Sentences are tokenised up to 512 tokens. Optimisation follows standard practice \citep{Wolf2020}: batch size 4, learning rate $2{\times}10^{-5}$, up to 10 epochs, weight decay 0.15, gradient accumulation 4, dropout 0.1, and early stopping with patience 4 epochs. We do not use class weights. All runs fit within an 8\,GB VRAM budget on a single GPU.

\paragraph{Hyperparameter selection}
We tuned the main hyperparameters for DeBERTa-base with Bayesian optimisation using Optuna \citep{Akiba2019}. For each configuration we ran 50 trials on the train/validation split, exploring:
number of epochs (3--10), batch size (2 or 4), learning rate ($[5\times10^{-6},\,5\times10^{-5}]$, log-uniform), and weight decay ($[0.1,\,0.3]$). We then retrained the final models from scratch using the best trial and the reference seed (42); the main configurations were subsequently re-run over three seeds (42, 7, and 1701), as described in Section~\ref{sec:thresholding}. The settings reported above correspond to the selected configuration.

\subsection{Lightweight auxiliary signals}
\label{sec:signals}

To stay within 8\,GB VRAM, we focus on simple, precomputed signals that can be fused with the DeBERTa representation at low cost. In all variants, auxiliary features are concatenated to the pooled sentence embedding and passed through a small projection layer before the final linear head, so that the backbone and optimisation settings remain identical across models.

\begin{enumerate}[leftmargin=1.25em,label=\alph*)]
  \item \textbf{Prior-sentence context and labels.}
  For a sentence $s$ occurring in a document, we concatenate up to the previous two sentences to the current sentence, separated by \texttt{[SEP]} tokens. This gives the text encoder a short-range view of the local discourse. In addition, we attach a vector that encodes the recent value activations.

  Let $\mathcal{V}$ be the set of 19 refined Schwartz values defined in Section~\ref{sec:task-definition}. For each sentence $s$ we form binary value vectors $\mathbf{y}_{s-1}, \mathbf{y}_{s-2} \in \{0,1\}^{|\mathcal{V}|}$ for the previous one and two sentences (zero vectors if the document is shorter). We then construct
  \[
  \mathbf{r}_s = \left[\, \mathbf{y}_{s-1} \;\|\; \mathbf{y}_{s-2} \right] \in \{0,1\}^{2|\mathcal{V}|},
  \]
  where $\|\,$ denotes concatenation and $|\mathcal{V}|$ is the number of refined Schwartz values (here $|\mathcal{V}| = 19$).

  To avoid any train--validation--test leakage, we implement these features in two stages. First, we train the direct DeBERTa value model described in Section~\ref{sec:direct} on the \emph{training} split only and then freeze its parameters. Second, we apply this frozen model once to all sentences in the train/validation/test splits and store its out-of-sample predictions $\hat{\mathbf{y}}_{s}$.

  When constructing $\mathbf{r}_s$ for the context-augmented models, we therefore always use:
  (i) the gold labels of $\mathbf{y}_{s-1}$ and $\mathbf{y}_{s-2}$ for sentences in the \emph{training} split, and
  (ii) the fixed out-of-sample predictions $\hat{\mathbf{y}}_{s-1}$ and $\hat{\mathbf{y}}_{s-2}$ produced by the frozen direct model for sentences in the \emph{validation} and \emph{test} splits.
  In particular, we never feed validation or test gold labels as inputs to any model; the test split is only used at evaluation time.

  This design removes any possibility of label leakage, but it introduces an asymmetry between training and inference that should be stated plainly. During training the model observes \emph{gold} prior-sentence labels, whereas at validation and test time it observes \emph{predicted} ones, which on this task are considerably noisier (the frozen direct model itself reaches macro-$F_1 \approx 0.32$). The feature is, therefore, learned under a form of teacher forcing and evaluated without it. A model that learns to rely on clean prior-sentence labels will find that signal degraded at inference, so any apparent benefit of this feature is better read as an upper bound than as a deployment estimate. Section~\ref{sec:results-lightweight} reports the empirical consequence: the feature adds $+0.002$ macro-$F_1$ over the calibrated baseline, a difference that a paired per-seed test cannot separate from seed variance, and we accordingly do not count it among this paper's findings. An alternative design would train on the frozen model's predictions as well, making the two regimes symmetric; we did not run that variant, and we flag it as the appropriate way to settle whether the feature has any real value.

  The vector $\mathbf{r}_s$ is passed through a learned linear projection to 16 dimensions with a ReLU non-linearity and concatenated to the pooled DeBERTa representation. This branch is intended to capture short-range discourse patterns and recently mentioned values without substantially increasing sequence length or VRAM usage.

  \item \textbf{Psycholinguistic and moral lexica.}
  We derive sentence-level feature vectors from several established lexica that have been widely used in affective, moral, and political NLP. Concretely, we use:
  \begin{itemize}
    \item \textbf{LIWC-22} (English dictionary; psychological and linguistic categories) \citep{Pennebaker2015};
    \item the \textbf{extended Moral Foundations Dictionary (eMFD)} \citep{Hopp2021}, which assigns words to Moral Foundations Theory categories (care/harm, fairness/cheating, etc.) with virtue/vice polarity;
    \item the \textbf{Moral Justification Dictionary (MJD)} \citep{WheelerLaham2016}, which captures variation in moral \emph{justification} content---deontological, consequentialist, and emotive language---across moral violations from different foundations;
    \item the \textbf{Schwartz value lexicon} released with the ValuesML / ValueEval resources \citep{ValueEval24Zenodo}, mapping words and short phrases to refined Schwartz values; and
    \item general \textbf{affective lexica}, including NRC VAD \citep{MohammadVAD2018}, NRC EmoLex \citep{MohammadTurney2013}, NRC Emotion Intensity \citep{Mohammad2018Intensity}, and the WorryWords lexicon \citep{Mohammad2024WorryWords}, which provide continuous scores for valence, arousal, dominance, and basic emotions.
  \end{itemize}

  For each lexicon $L$ with category set $\mathcal{C}_L$ and a sentence $s$ tokenised into words $\{w_1,\dots,w_n\}$, we compute a raw sentence-level feature vector $\mathbf{f}_L(s) \in \mathbb{R}^{|\mathcal{C}_L|}$ as follows.

  \begin{itemize}
    \item For lexica that assign binary membership of words to categories (e.g., LIWC-22, eMFD, MJD, Schwartz), we use \emph{length-normalised category frequencies}:
    \[
      f_{L,c}(s) \;=\; \frac{1}{n} \sum_{i=1}^{n} \mathbb{I}[w_i \in L_c],
    \]
    where $L_c$ is the set of words associated with category $c \in \mathcal{C}_L$.
    \item For lexica that provide continuous scores (e.g., NRC VAD, NRC EmoLex), we average the scores over all tokens in $s$ that are present in the lexicon and set the value to zero when no token in $s$ is found in that category. This yields a dense, length-normalised vector over the lexicon dimensions.
  \end{itemize}

  The resulting vectors are sparse and relatively low-dimensional compared to the DeBERTa representation. To obtain a compact representation with uniform size across lexica, we apply a separate learned linear projection for each lexicon:
  \[
    \mathbf{h}_L(s) = \sigma\!\left( W_L \mathbf{f}_L(s) + \mathbf{b}_L \right),
  \]
  where $W_L \in \mathbb{R}^{128 \times |\mathcal{C}_L|}$, $\mathbf{b}_L \in \mathbb{R}^{128}$, and $\sigma$ is a ReLU activation. We set the projection size to 128 dimensions as a compromise between capacity and efficiency: it is large enough to capture interactions between lexicon categories, yet small relative to the 768-dimensional DeBERTa-base sentence embedding, so that the additional parameters and VRAM cost remain modest. Intuitively, this branch allows the model to attend explicitly to affective, stylistic, and moral cues that might be underrepresented in the purely contextual representation, which is particularly helpful for rare values.

  \item \textbf{Topics.}
  To provide coarse issue and domain information, we add topic indicators from three unsupervised models trained on the training split: LDA with 60 topics, NMF with 90 topics, and BERTopic \citep{Jelodar2019,LeeSeung1999,Grootendorst2022}. Each model yields, for a sentence $s$, a probability vector over its topics (e.g., a 60-dimensional distribution for LDA). We treat these probability vectors as continuous features, apply a learned linear projection to 128 dimensions with a ReLU activation (analogous to the lexicon branch), and concatenate the resulting topic embeddings with the text representation. These topic features capture high-level issues (e.g., security, environment, economy) that can disambiguate otherwise neutral sentences and reduce the burden on the encoder to recover global context from a single sentence.
\end{enumerate}

Let $\mathbf{h}_{\text{text}}(s)$ denote the pooled DeBERTa representation for sentence $s$, and let $\mathbf{h}_{\text{prior}}(s)$, $\mathbf{h}_{\text{lex}}(s)$, and $\mathbf{h}_{\text{topic}}(s)$ be the concatenation of the projected prior-sentence, lexicon, and topic embeddings that are active in a given configuration. The final fused representation is
\[
  \mathbf{h}_{\text{fused}}(s) = \left[\, \mathbf{h}_{\text{text}}(s) \;\|\; \mathbf{h}_{\text{prior}}(s) \;\|\; \mathbf{h}_{\text{lex}}(s) \;\|\; \mathbf{h}_{\text{topic}}(s) \right],
\]
which is then passed through dropout and a linear prediction head. This fusion architecture is shared across all feature-augmented variants; ablations that remove or combine branches are reported in the Supplementary Material.

\subsection{Presence gate and hierarchical pipeline}
\label{sec:hierarchy}

We implement a two-stage hierarchy:

\begin{enumerate}[leftmargin=1.25em,label=\arabic*)]
  \item \textbf{Presence gate.} A binary classifier predicts $z_s$ (``any value present?''). We use two variants:
  (i) a supervised DeBERTa-based presence model (with or without auxiliary features as in Section~\ref{sec:signals}), and
  (ii) a lightweight SBERT+logistic model, where \texttt{all-MiniLM-L6-v2} sentence embeddings \citep{ReimersGurevych2019SBERT,Wang2020MiniLM} feed a class-weighted logistic regression. The gate outputs a probability $g_s \in [0,1]$.
  \item \textbf{Value head.} A 19-label classifier predicts $\{\hat{p}_{s,v}\}_{v\in\mathcal{V}}$. In the hierarchical setup we apply a hard mask: if $g_s < \tau_{\text{gate}}$ we set $\hat{p}_{s,v} \leftarrow 0$ for all $v$, otherwise we keep $\hat{p}_{s,v}$ as predicted.
\end{enumerate}

We tune $\tau_{\text{gate}}$ on the validation set to maximise end-to-end macro-$F_1$ on value prediction, not gate performance alone.

In our implementation the presence gate and the value classifier are trained as two separate models. Both use a DeBERTa-base backbone and the same optimisation protocol: the gate is trained on the binary labels $z_s$, while the hierarchical value classifier is trained on the 19-dimensional vectors $\{y_{s,v}\}_{v\in\mathcal{V}}$ \emph{restricted to the presence-positive sentences} (those with at least one active value). The direct value classifier, in contrast, is trained on all sentences. During evaluation of the hierarchical variants (RQ2), we obtain gate probabilities $g_s$ and value probabilities $\hat{p}_{s,v}$ on the test set and apply a hard mask with a validation-tuned gate threshold $\tau_{\text{gate}}$ as described above. Direct and hierarchical models therefore differ in two respects: their training distribution (all sentences vs.\ presence-positive sentences) and the inference-time gate. Because this leaves the direct-vs-gated comparison confounded by training distribution as well as architecture, we additionally report a feature-matched contrast (direct vs.\ presence-gated MJD, sharing the same features) in Section~\ref{sec:results-hierarchy}.

\subsection{Instruction-tuned LLM baselines and QLoRA}
\label{sec:llms}

We benchmark instruction-tuned open LLMs that can be run on a single mid-range GPU (8\,GB VRAM in our setup), specifically Llama~3.1~8B, Gemma~2~9B, Ministral~8B, and Qwen~2.5~7B (HuggingFace checkpoints \texttt{meta-llama/\allowbreak Llama-3.1-8B-Instruct}, \texttt{google/\allowbreak gemma-2-9b-it}, \texttt{mistralai/\allowbreak Ministral-8B-\allowbreak Instruct-2410}, and \texttt{Qwen/\allowbreak Qwen2.5-7B-Instruct}, respectively). All LLM experiments are conducted using the HuggingFace transformers stack with 4-bit NF4 quantisation via \texttt{bitsandbytes}, and, in the QLoRA setups, low-rank adapters are trained on top of the frozen 4-bit base models. We do not rely on specialised inference frameworks such as LightLLM or \texttt{ollama}; instead, all models are run in this standard 4-bit configuration under the same single-GPU constraint. We use three settings:

\begin{itemize}[leftmargin=1.5em]
  \item \textbf{Zero-shot prompting.} The model receives a sentence and natural-language definitions of the 19 values and is asked to return a JSON array of active values.
  \item \textbf{Few-shot prompting.} We prepend $k$ labelled examples ($k \in \{1,2,4,8,\allowbreak 12,16,20\}$) that illustrate both positive and negative cases per value, and use the same JSON-style output format.
  \item \textbf{QLoRA fine-tuning.} We fine-tune the best-performing LLM (Gemma~2 9B) with quantised low-rank adaptation \citep{Dettmers2024,Hu2022} on the training split. The ``QLoRA direct'' configuration uses rank $r=16$, $\alpha=32$, three epochs, gradient accumulation 8, maximum sequence length 512, and learning rate $2\times 10^{-4}$. The ``QLoRA hier (SBERT gate)'' variant uses $r=8$, $\alpha=16$, three epochs, maximum length 256, and the same learning rate. Adapters are applied to the $\{\texttt{q\_proj,k\_proj,v\_proj,o\_proj}\}$ modules and only adapters are saved.
\end{itemize}

Generation uses greedy decoding (no sampling), with at most 200 new tokens (\texttt{max\_new\_tokens}). We post-process model outputs to obtain a binary 19-dimensional vector. Decision thresholds for LLM probabilities are tuned on validation and then fixed for test.

We also evaluate a retrofit hierarchy where an SBERT-based presence gate (Section~\ref{sec:hierarchy}) zeroes out LLM probabilities for $g_s < \tau_{\text{gate}}$, with $\tau_{\text{gate}}$ tuned on validation for end-to-end macro-$F_1$.

Larger proprietary models (e.g., GPT-4-class systems) are outside the scope of this study: our aim is to compare architectures that can be deployed in typical academic or practitioner settings with limited GPU resources.

\subsection{Ensembles}
\label{sec:ensembles}

We form compact ensembles by forward selection over a pool of candidate runs. We consider:

\begin{itemize}[leftmargin=1.5em]
  \item \textbf{Hard voting} on binarised predictions.
  \item \textbf{Soft or weighted voting} on probabilities, with weights proportional to validation macro-$F_1$.
\end{itemize}

For soft voting we compute the averaged (or weighted) probabilities per value and select a global threshold $t^\star$ on validation by sweeping $t \in [0,1]$ (step 0.01). The chosen $t^\star$ is then applied unchanged to the test set.

Forward selection proceeds greedily: starting from the best single model, we add candidates one by one and keep a new candidate only if the one-sided bootstrap lower 95\% confidence bound for the improvement in macro-$F_1$ is both $>\!0$ in absolute terms and at least $1\%$ in relative terms. We build separate ensembles for (i) direct DeBERTa models, (ii) presence-gated models, and (iii) LLM-based systems, as well as mixed ensembles that combine models from different families.

\subsection{Statistical testing and reporting}
\label{sec:significance}

We use nonparametric tests for all paired model comparisons.

\paragraph{Bootstrap tests}
For macro-$F_1$ comparisons we draw $B{=}2000$ bootstrap samples over instances \citep{Efron1979}. For each pair of systems we estimate the distribution of $\Delta \text{Macro-}F_{1}$, report its mean, a one-sided lower 95\% confidence bound, and a one-sided $p$-value for the hypothesis that the more complex model does not improve over the simpler one.

\paragraph{McNemar tests}
For per-value analysis we apply McNemar's exact test \citep{McNemar1947} to the positive class of each value. We control the False Discovery Rate at $\alpha=0.05$ using the Benjamini--Hochberg procedure \citep{BenjaminiHochberg1995}.

\paragraph{Paired per-seed tests}
Bootstrap resampling over instances estimates sampling noise with the training seed held fixed, and therefore cannot establish that a difference exceeds \emph{seed} variance. We consequently add a seed-level analysis: for each contrast we pair the two configurations within every seed, each evaluated at its own validation-tuned threshold, and test the resulting per-seed differences with a paired $t$-test. We report the mean difference, whether it keeps the same sign across all three seeds, and its magnitude relative to the seed-to-seed standard deviation of the configurations being compared. With three seeds this test has two degrees of freedom and correspondingly low power, so we treat sign consistency and the ratio of the effect to seed noise as the primary evidence, and the $p$-value as secondary.

For transparency, we summarise the main significance comparisons in tables. Section~S7 in the Supplementary Material reports, for the key model pairs discussed in Section~\ref{sec:results}, their test macro-$F_1$, the bootstrap estimate of $\Delta \text{Macro-}F_{1}$, the one-sided lower 95\% confidence bound, and the corresponding one-sided $p$-value. Section~S7 also lists those values for which the difference between two systems on the positive class remains significant after Benjamini--Hochberg correction.

\section{Experimental setup}
\label{sec:experimental-setup}

We evaluate all models on the English, machine-translated ValueEval'24 splits described in Section~\ref{sec:dataset} and summarised in Section~\ref{sec:stats}. We use the official train/validation/test partition without additional filtering. As defined in Section~\ref{sec:task-definition}, we consider two sentence-level tasks: (i) binary detection of \texttt{presence} (at least one value active), and (ii) 19-way multi-label prediction of the refined Schwartz values.

\subsection{Models: direct vs.\ hierarchical variants}
\label{sec:direct-vs-hierarchical}

For both tasks we reuse the model families introduced in Section~\ref{sec:methods}:

\begin{itemize}[leftmargin=1.5em]
  \item \textbf{Direct models} are single-branch DeBERTa-base classifiers that map a sentence $s$ to either a single probability $\hat{p}_s$ (\texttt{presence}) or a 19-dimensional vector $\{\hat{p}_{s,v}\}_{v\in\mathcal{V}}$ (values). Some variants include lightweight auxiliary signals (context, lexica, topics) as described in Section~\ref{sec:signals}.
  \item \textbf{Hierarchical models} add a binary \emph{presence gate} in front of the value detector (Section~\ref{sec:hierarchy}). The gate predicts $g_s \in [0,1]$ (``any value?''). If $g_s < \tau_{\text{gate}}$, all value probabilities are set to zero; otherwise the value model outputs are kept. We experiment with DeBERTa-based gates and a lightweight SBERT+logistic alternative.
  \item \textbf{LLM-based models} (Section~\ref{sec:llms}) use instruction-tuned 7--9B open models (Gemma~2~9B, Llama~3.1~8B, Ministral~8B, Qwen~2.5~7B) via zero-/few-shot prompting or QLoRA. In the hierarchical variants, an SBERT presence gate filters the LLM predictions.
  \item \textbf{Ensembles} (Section~\ref{sec:ensembles}) combine a small number of these models via hard or soft voting.
\end{itemize}

Unless explicitly stated, all architectures and feature branches are exactly those described in Section~\ref{sec:methods}; the role of this section is to clarify training, thresholding, and evaluation choices.

\subsection{Training protocol and hyperparameters}
\label{sec:training-protocol}

All transformer-based models share the optimisation setup in Section~\ref{sec:direct}. Sentences are tokenised with the DeBERTa-base tokenizer and truncated or padded to 512 WordPiece embeddings (subword tokens produced by the WordPiece-style tokenizer; cf.\ \citep{Devlin2019}). We train with AdamW, batch size 4 (gradient accumulation 4, effective batch size 16), learning rate $2\times10^{-5}$, weight decay 0.15, dropout 0.1, and up to 10 epochs with early stopping on validation macro-$F_1$ (patience 4). The best validation checkpoint is used for test evaluation.

Note that gradient accumulation does not increase the peak VRAM usage beyond that of a single micro-batch. In our setup, a per-device batch size of 4 sentences is processed at a time and gradients are accumulated across 4 such micro-batches before each optimiser step, which yields an effective batch size of 16 without ever holding 16 sentences simultaneously on the GPU. Measured with \texttt{nvidia-smi} on an NVIDIA GeForce RTX~3070 (8\,GB VRAM), the peak allocated memory during DeBERTa-base fine-tuning with sequence length 512, batch size 4 and gradient accumulation 4 was approximately 7.5\,GB. All reported DeBERTa experiments respect this per-step memory budget.

We do not use class weights for \texttt{presence}, as the label is roughly balanced across splits (Section~\ref{sec:stats}), and we keep the same optimisation settings for text-only and feature-augmented variants. All DeBERTa configurations are trained with three random seeds (42, 7, and 1701), and results are reported as the mean over these seeds with per-seed standard deviations (Section~\ref{sec:thresholding}); seed~42 serves as the reference seed for quantities reported at a single seed, such as the tuned thresholds shown in Tables~\ref{tab:direct-vs-hier} and~\ref{tab:lightweight-signals}. For architectures with auxiliary branches (Section~\ref{sec:signals}), we perform a small grid search on the validation set over dropout rate $\{0.1,0.2\}$ and auxiliary projection size (64 vs.\ 128 for lexicon/topic branches) and reuse the best setting across comparable models to limit compute.

For all variants that use previous-sentence label vectors as auxiliary features (Section~\ref{sec:signals}), we follow the two-stage procedure outlined there to avoid information leakage. Concretely, we first train the direct DeBERTa value model on the training split only and freeze it. We then apply this frozen model once to all sentences in the train/validation/test splits and store its predictions, which are treated as fixed features for the context-augmented models. These models are subsequently trained only on the training split, with early stopping and all threshold tuning based exclusively on the validation split. The test split is used exactly once for the final evaluation and is never involved in training, hyperparameter selection, or feature construction beyond the one-off forward pass of the frozen direct model. As discussed in Section~\ref{sec:signals}, this procedure is leakage-free but asymmetric: prior-sentence label features are built from gold labels during training and from predicted labels at validation and test time.

LLM-based models are run and fine-tuned as described in Section~\ref{sec:llms}. All LLM configurations (zero-/few-shot and QLoRA) are trained and evaluated on the same ValueEval'24 splits as the DeBERTa models.

\subsection{Thresholding and evaluation metrics}
\label{sec:thresholding}

For the multi-label value task, models output probabilities $\hat{p}_{s,v}$ for each value $v\in\mathcal{V}$. We convert them to binary predictions $\hat{y}_{s,v}$ using the decision rules in Section~\ref{sec:notation-thresholds}: either a fixed global threshold ($0.5$) or label-wise tuned thresholds $\tau_v$ selected on the validation set. Our primary metric is macro-averaged $F_1$ over the positive class across the 19 values.

For the \texttt{presence} task, models output a probability $\hat{p}_s$ for \texttt{presence}$=1$. We consider:

\begin{itemize}[leftmargin=1.5em]
  \item a fixed global threshold $t=0.5$; and
  \item a tuned threshold $t^\star$, selected on the validation set by sweeping $t\in[0.10,0.90)$ in steps of 0.01 and choosing the value that maximises positive-class recall subject to a minimum precision of $0.40$ (the label-wise scheme of Section~\ref{sec:notation-thresholds}).
\end{itemize}

The tuned threshold $t^\star$ is then applied unchanged to the test set. The main evaluation metric for presence is again the positive-class $F_1$ score; we also monitor accuracy and AUC on validation to detect overfitting, but we do not optimise directly for them.

Statistical significance for all paired comparisons (presence and values) follows the bootstrap and McNemar protocol in Section~\ref{sec:significance}. Unless otherwise noted, we report scores rounded to three decimal places; multi-seed results (Sections~\ref{sec:results-presence}--\ref{sec:results-lightweight}) are reported as the mean over seeds 42, 7, and 1701, with per-seed standard deviations on the test columns.

\subsection{Hardware and software environment}
\label{sec:hardware-software}

All experiments run on a single NVIDIA GPU with 8\,GB of VRAM (GeForce RTX~3070), a commodity CPU, and 12--16\,GB of host RAM. This single-GPU constraint determines the choice of backbone models, batch sizes, and LLM sizes. For encoder-based models (DeBERTa-base and variants), we empirically measured peak allocated memory with \texttt{nvidia-smi}: with maximum sequence length 512, per-device batch size 4, and gradient accumulation 4, peak VRAM usage during training was approximately 7.5\,GB. During inference, DeBERTa-based models were run with batch size $\leq 4$, resulting in peak VRAM usage below 7\,GB.

For instruction-tuned LLMs (Gemma~2~9B, Llama~3.1~8B, Ministral~8B, Qwen~2.5~7B), we always load the base model weights in 4-bit NF4 quantisation using \texttt{bitsandbytes} and apply parameter-efficient adapters via QLoRA. Fine-tuning is performed with batch size 1 and gradient accumulation, and peak VRAM usage for both training and inference remains below 8\,GB in all configurations. We deliberately avoid techniques such as tensor parallelism or model offloading to larger GPUs, so that all reported results correspond to setups that can be reproduced on a modest single-GPU budget.

We use Python~3.10, PyTorch, and \texttt{transformers} versions from early 2024; exact package versions, configuration files, and tuned thresholds are released with the code to enable replication. The hardware and software environment is shared across all model families.

\section{Results and discussion}
\label{sec:results}

We organize the results around the four research questions introduced in Section~\ref{sec:introduction}: (RQ1) feasibility of detecting the \emph{presence} of moral content, (RQ2) hierarchical vs.\ direct value detection, (RQ3) impact of lightweight signals, and (RQ4) comparison between supervised DeBERTa models and instruction-tuned LLMs and their ensembles.

\subsection{RQ1: Can we reliably detect the presence of moral content in single sentences?}
\label{sec:results-presence}

We first treat \texttt{presence} as a binary label (1 iff at least one of the 19 values is positive; Section~\ref{sec:task-data}) and train DeBERTa-base classifiers that only predict this gate. All models share the same backbone and training protocol (Section~\ref{sec:experimental-setup}) and differ only in the auxiliary features concatenated to the sentence representation.

Table~\ref{tab:presence-gate} reports the strongest presence-gate configurations; all figures are means over three seeds (42, 7, 1701). The text-only baseline already achieves a validation positive-class $F_1$ of $0.73$ and a test positive-class $F_1$ of $0.73$ at threshold $t=0.5$, essentially unchanged when we tune the threshold on validation (best $t^\star = 0.10$). Adding LIWC-22 features, or combining LIWC-22, EmoLex, or eMFD with the previous two sentences (plus their labels), yields comparable performance: all strong variants reach test positive-class $F_1$ of about $0.72$--$0.73$ at both $t=0.5$ and $t^\star=0.10$, with per-seed standard deviation $\le 0.014$.

\begin{table*}[t]
\caption{Binary \texttt{presence} detection on the English splits. All $F_1$ values (validation and test) are positive-class $F_1$. For tuned thresholds we sweep $t$ on validation and choose the $t^\star$ that maximises positive-class recall subject to a minimum precision of $0.40$ (the label-wise scheme of Section~\ref{sec:notation-thresholds}), then apply $t^\star$ unchanged to test. Each entry is the mean over three seeds (42, 7, 1701); test columns additionally report the sample standard deviation, which is $\le 0.014$ throughout. The tuned threshold is $t^\star = 0.10$ for every row.}
\centering
\small
\setlength{\tabcolsep}{4pt}
\begin{tabular}{p{.16\linewidth} p{.23\linewidth} c c c}
\toprule
\textbf{Presence model} & \textbf{Aux.\ features} & \textbf{Val $F_1$} & \textbf{Test $F_1$ @ $0.5$} & \textbf{Test $F_1$ @ $t^\star$} \\
\midrule
Baseline (text only) & -- & 0.729 & $0.728 \pm 0.013$ & $0.732 \pm 0.005$ \\
LIWC-22 + linguistic & LIWC-22 + ling.\ cats. & 0.725 & $0.724 \pm 0.012$ & $0.732 \pm 0.003$ \\
Prev-2 + LIWC-22 & 2 prev.\ sent.\ + labels + LIWC-22 & 0.731 & $0.730 \pm 0.014$ & $0.730 \pm 0.011$ \\
Prev-2 + EmoLex & 2 prev.\ sent.\ + labels + EmoLex & 0.720 & $0.721 \pm 0.006$ & $0.732 \pm 0.004$ \\
Prev-2 + eMFD & 2 prev.\ sent.\ + labels + eMFD & 0.715 & $0.719 \pm 0.014$ & $0.730 \pm 0.007$ \\
\bottomrule
\end{tabular}
\label{tab:presence-gate}
\end{table*}

Overall, RQ1 is answered positively: \emph{moral presence is reliably learnable from single sentences}. Several gate configurations achieve $F_1 \approx 0.73$ on the test set, and differences between strong variants are small compared to this ceiling. In the remainder of the paper, presence is used either as a stand-alone signal (for filtering) or as a gate in the hierarchical architectures for the 19-value task; it is never used as a replacement for the fine-grained value labels themselves.

\subsection{RQ2: Does a moral-presence gate help over a direct 19-way value detector under matched compute?}
\label{sec:results-hierarchy}

We next compare a direct multi-label value detector with a hierarchical pipeline that first predicts \texttt{presence} and only runs the value classifier on sentences predicted as moral. In all cases, the value classifier is trained in exactly the same way as in the direct setting; the only difference is that, at test time, its 19-dimensional output is multiplied by a binary mask derived from the independently trained presence gate. Both setups use DeBERTa-base; the gate and the value classifier are trained as in Section~\ref{sec:experimental-setup}. We focus on the best-performing variants in Table~\ref{tab:direct-vs-hier}; full results for all hierarchical configurations appear in Section~S4 in the Supplementary Material.

As a direct baseline we use the text-only DeBERTa model with a tuned global threshold (reference $t^\star = 0.30$) applied to all 19 labels, which obtains $F_1 = 0.315 \pm 0.004$ on the test set (mean $\pm$ std over three seeds). Adding LIWC-22 features and tuning the threshold improves this to $F_1 = 0.320 \pm 0.003$ ($t^\star = 0.25$), making it our best \emph{single} direct Transformer. For the hierarchical setting we select the best-performing presence-gated configuration on validation: a value classifier enriched with the Moral Justification Dictionary (MJD), gated by a strong presence model that uses the previous two sentences plus LIWC-22 features at $t_{\text{gate}} = 0.10$. This pipeline reaches $F_1 = 0.302 \pm 0.009$ on test with a tuned value threshold (reference $t^\star = 0.40$). To separate the effect of the gate from that of the training distribution, we also train a \emph{direct} MJD head on all sentences with the same MJD features; this feature-matched direct model reaches $F_1 = 0.316 \pm 0.001$, above the gated variant.

\begin{table*}[t]
\caption{Direct vs.\ hierarchical DeBERTa value detectors on the test set (macro $F_1$ over 19 values). All models share the same backbone and training protocol; only feature sets and the presence gate differ. The \emph{Direct + MJD} and \emph{Presence-gated (MJD)} rows form a feature-matched contrast---identical MJD features, differing only in training distribution (all sentences vs.\ presence-positive only) and the inference-time gate. Values are mean $\pm$ std over three seeds (42, 7, 1701); thresholds are tuned per seed and the reference-seed value is shown.}
\centering
\small
\setlength{\tabcolsep}{4pt}
\begin{tabular}{>{\raggedright\arraybackslash}p{.15\linewidth} >{\raggedright\arraybackslash}p{.18\linewidth} >{\raggedright\arraybackslash}p{.20\linewidth} >{\raggedright\arraybackslash}p{.18\linewidth} c}
\toprule
\textbf{Model} & \textbf{Architecture} & \textbf{Features} & \textbf{Threshold(s)} & \textbf{Test $F_1$} \\
\midrule
Direct baseline & Direct & Text only & $t = 0.30$ & $0.315 \pm 0.004$ \\
Direct + LIWC-22 & Direct & Text + LIWC-22 & $t = 0.25$ & $\mathbf{0.320 \pm 0.003}$ \\
Direct + MJD & Direct & Text + MJD & $t = 0.25$ & $0.316 \pm 0.001$ \\
Presence-gated (MJD) & Hierarchical & Text + MJD, gate = Prev-2 + LIWC-22 & $t = 0.40$, $t_{\text{gate}} = 0.10$ & $0.302 \pm 0.009$ \\
\bottomrule
\end{tabular}
\label{tab:direct-vs-hier}
\end{table*}

Across the three seeds, the presence-gated pipeline (mean macro-$F_1 = 0.302 \pm 0.009$) does not exceed the direct baseline ($0.315 \pm 0.004$) and stays clearly below the best direct LIWC-22 model ($0.320 \pm 0.003$) under the same 8\,GB budget; the feature-matched contrast (direct vs.\ presence-gated MJD, $0.316 \pm 0.001$ vs.\ $0.302 \pm 0.009$) points the same way. In our setting, \emph{presence gating does not yield a macro-$F_1$ gain over direct value detection}, and gate recall becomes a bottleneck.

\subsection{RQ3: Which lightweight auxiliary signals help under a fixed single-GPU compute budget?}
\label{sec:results-lightweight}

We now ablate the lightweight signals introduced in Section~\ref{sec:signals} on the direct DeBERTa value detector. Table~\ref{tab:lightweight-signals} shows representative configurations. All models share the same backbone, training protocol, and decision-threshold tuning strategy; the only differences are the auxiliary features and the tuned global threshold $t^\star$.

The single largest lever in this ablation is decision-threshold calibration. Moving the text-only baseline from the default $t=0.5$ ($F_1=0.282$) to a tuned global threshold ($F_1=0.315$) yields a $+0.033$ macro-$F_1$ improvement that is consistent across all three seeds (paired per-seed $t$-test, $p = 0.004$; $12$ times the baseline's seed-to-seed standard deviation)---on its own larger than the effect of any individual auxiliary signal, and enough to bring a standard baseline to or above the best official ValueEval'24 English run ($\approx 0.28$). This is a direct consequence of the severe class imbalance: at $t=0.5$ the model systematically under-predicts rare values, and threshold tuning rebalances precision and recall across the 19 labels. On top of this calibrated baseline, simple features add far less: short-range context (previous two sentences and their labels) reaches $0.317$, LIWC-22 features $0.320$, and BERTopic topics $0.320$ (means over three seeds; per-seed std $\le 0.008$), i.e.\ mean gains of only $+0.002$ to $+0.005$ macro-$F_1$, which a paired per-seed test does not separate from seed variance. Token pruning, deeper classifier heads (two-layer MLP, residual block), and naive augmentation did not consistently improve validation performance and were not pursued further.

\paragraph{Does the calibration effect hold for another encoder?}
Because this is the paper's central empirical claim, we tested whether it is a property of DeBERTa or of the task. We repeated the text-only experiment with \texttt{FacebookAI/\allowbreak roberta-base}, holding everything else fixed: the same corpus and official splits, the same optimisation settings (Section~\ref{sec:training-protocol}), the same three seeds, and the same validation-tuned global threshold procedure. The encoder is the only thing that differs. Table~\ref{tab:cross-encoder} gives the comparison.

The effect replicates, and is larger for the weaker encoder. RoBERTa-base is slightly behind DeBERTa-base once calibrated ($0.309$ vs.\ $0.315$), consistent with the model-selection comparison reported for this corpus \citep{Yeste2024}. Calibration lifts it by $+0.043$ against $+0.033$ for DeBERTa, with the same sign in all three seeds for both encoders and roughly eight times the seed-to-seed standard deviation in each case. The RoBERTa gain is noisier ($\pm 0.014$ against $\pm 0.004$) and its paired $p$ weaker ($0.033$ against $0.004$): significant, though less decisively.

Threshold calibration is thus not a quirk of one architecture. It addresses a property of the decision rule under severe class imbalance, the systematic under-prediction of rare labels at $t = 0.5$, which is largely independent of how the encoder represents the sentence. We state the boundary of this evidence: two encoder families on one corpus is a replication, not a demonstration of universality.

\begin{table*}[t]
\caption{Effect of lightweight signals on direct DeBERTa value detection (macro $F_1$ over 19 values). All models use the same backbone and hyperparameters; only features and the global decision threshold $t^\star$ differ. Test columns are mean $\pm$ std over three seeds (42, 7, 1701); the validation column is the seed-mean at $t=0.5$, and $t^\star$ is tuned per seed with the reference-seed value shown (it varies by at most one grid step across seeds).}
\centering
\small
\setlength{\tabcolsep}{4pt}
\fitwidth{%
\begin{tabular}{p{.15\linewidth} p{.20\linewidth} c c c c}
\toprule
\textbf{Direct model} & \textbf{Aux.\ features} & \textbf{Val $F_1$} & \textbf{Test @ $0.5$} & \textbf{Test @ $t^\star$} & \textbf{$t^\star$} \\
\midrule
Baseline (text only) & -- & 0.279 & $0.282 \pm 0.001$ & $0.315 \pm 0.004$ & 0.30 \\
Prev-2 sentences + labels & Context & 0.277 & $0.280 \pm 0.008$ & $0.317 \pm 0.001$ & 0.15 \\
LIWC-22 & Psycholing.\ lexicon & 0.284 & $0.288 \pm 0.004$ & $\mathbf{0.320 \pm 0.003}$ & 0.25 \\
BERTopic topics & Topic distribution & 0.279 & $0.284 \pm 0.002$ & $0.320 \pm 0.001$ & 0.20 \\
\bottomrule
\end{tabular}}
\label{tab:lightweight-signals}
\end{table*}

\begin{table*}[t]
\caption{Threshold calibration on two encoder families, text-only, under an otherwise identical protocol (mean $\pm$ std over seeds 42, 7, 1701; each seed's threshold tuned on validation and applied unchanged to test). The calibration gain is computed per seed and then averaged, and is tested with the paired per-seed procedure of Section~\ref{sec:significance}.}
\centering
\small
\setlength{\tabcolsep}{6pt}
\fitwidth{%
\begin{tabular}{l c c c c}
\toprule
\textbf{Encoder} & \textbf{Test $F_1$ @ $0.5$} & \textbf{Test $F_1$ @ $t^\star$} & \textbf{Calibration gain} & \textbf{Paired $p$} \\
\midrule
DeBERTa-base & $0.282 \pm 0.001$ & $0.315 \pm 0.004$ & $+0.033 \pm 0.004$ & $0.004$ \\
RoBERTa-base & $0.267 \pm 0.009$ & $0.309 \pm 0.005$ & $+0.043 \pm 0.014$ & $0.033$ \\
\bottomrule
\end{tabular}}
\label{tab:cross-encoder}
\end{table*}

These feature gains over the \emph{calibrated} baseline do not survive a paired per-seed test. Pairing each configuration against the baseline within each seed, with each at its own validation-tuned threshold (Section~\ref{sec:significance}), the mean differences are $+0.002$ for previous-sentence context, $+0.005$ for LIWC-22 and $+0.004$ for BERTopic topics; each reverses sign on at least one of the three seeds, and none approaches significance (paired $t$-test $p = 0.55$, $0.32$ and $0.22$ respectively, two degrees of freedom). The effects are of the same order as the seed-to-seed standard deviation of the calibrated baseline itself ($0.004$). We do not claim these auxiliary signals as established improvements: at three seeds, differences of this size cannot be separated from seed variance, and resolving them would require substantially more runs than our compute budget allows. The contrast with threshold calibration is the substantive point: calibration improves every seed and is roughly twelve times the baseline's seed standard deviation. That is why we recommend calibration, rather than feature engineering, as the first design lever.

\paragraph{Why the auxiliary signals add so little}
Two mechanisms explain this, and they reinforce each other. First, the failure mode that dominates macro-$F_1$ is systematic under-prediction of rare values at $t = 0.5$, which is a property of the decision rule rather than of the representation; calibration addresses it directly and across all 19 labels at once, whereas a lexical feature can only help where its vocabulary actually occurs. Second, and less obviously, the values that word-frequency features are best placed to help are the ones the encoder already handles well. As the per-value analysis in Section~\ref{sec:error-analysis} shows, the two best-detected values are precisely the lexically distinctive ones, \emph{Universalism: nature} and \emph{Tradition}, while the values that remain near the floor are abstract and diffusely expressed. Lexical resources have little headroom on the first group and little purchase on the second, which is why they cannot move the macro average appreciably.

The prior-sentence label feature deserves a specific caveat beyond this. As described in Section~\ref{sec:signals}, it is trained on gold prior-sentence labels but evaluated on predicted ones, so its $+0.002$ is measured under a train--inference asymmetry that would, if anything, flatter it. A feature that is already indistinguishable from seed noise under a protocol biased in its favour cannot be presented as a contribution, and we do not do so.

\subsection{RQ4: How do supervised DeBERTa models compare to instruction-tuned open LLMs and their ensembles?}
\label{sec:results-llms}

Finally, we compare the best supervised DeBERTa models to instruction-tuned LLMs (Gemma~2~9B, Llama~3.1~8B, Ministral~8B, Qwen~2.5~7B) in zero-shot, few-shot, and QLoRA setups, as well as their ensembles (Section~\ref{sec:llms}). Table~\ref{tab:transformers-vs-llms} summarises the strongest \emph{single} and \emph{ensemble} configurations for each family; full LLM and ensemble results appear in Section~S5 in the Supplementary Material.

Among LLMs, the best single configuration is Gemma~2~9B with few-shot prompting and an SBERT presence gate (definition-style prompting with 20 in-context examples and $t_{\text{gate}} = 0.29$), which reaches macro-$F_1 = 0.238$ on the test set. The same Gemma model without the gate obtains $F_1 = 0.233$; the gated variant is significantly better, but still well below supervised DeBERTa. QLoRA fine-tuning of Gemma in a direct configuration reaches $F_1 = 0.236$, close to the best few-shot result.

Ensembling LLMs helps: a hard-voting ensemble of the best gated few-shot Gemma and its QLoRA variant reaches $F_1 = 0.277$, a statistically significant gain over the best single LLM. However, supervised DeBERTa remains clearly ahead. The best single direct DeBERTa variant (with LIWC-22 features and tuned threshold) reaches $F_1 = 0.320$. A soft-voting ensemble of three direct DeBERTa models (baseline, LIWC-22, and Prev-2 sentences), with a tuned global threshold $t^\star = 0.29$, achieves our overall best result with $F_1 = 0.332$.

\begin{table*}[t]
\caption{Comparison between supervised DeBERTa and LLM-based systems on the test set (macro $F_1$ over 19 values). The reporting basis differs by row and is stated here for clarity: the single-transformer row is the three-seed mean of Table~\ref{tab:lightweight-signals} ($0.320 \pm 0.003$ over seeds 42, 7, 1701), whereas the ensemble and LLM rows are reference-seed (42) values, since the soft-voting ensemble is composed from the reference-seed component models and each LLM configuration is run once.}
\centering
\small
\begin{tabular}{p{.17\linewidth} p{.35\linewidth} l c}
\toprule
\textbf{Family} & \textbf{Configuration} & \textbf{Architecture} & \textbf{Test $F_1$} \\
\midrule
Transformer (single) & DeBERTa + LIWC-22, tuned & Direct & 0.320 \\
Transformer (ensemble) & Baseline + LIWC-22 + Prev-2, soft vote, $t^\star=0.29$ & Direct ensemble & \textbf{0.332} \\
LLM (single) & Gemma-2-9B few-shot + SBERT presence gate & Hier.\ (presence gate) & 0.238 \\
LLM (ensemble) & Gemma few-shot hier + Gemma QLoRA, hard vote & LLM ensemble & 0.277 \\
\bottomrule
\end{tabular}
\label{tab:transformers-vs-llms}
\end{table*}

Bootstrap tests confirm that the DeBERTa ensemble significantly improves both over its strongest single component (DeBERTa~+~LIWC-22: $\Delta\text{macro-}F_1 = +0.012$, one-sided lower 95\% bound $0.007$, $p<10^{-3}$) and over the best LLM ensemble ($\Delta\text{macro-}F_1 = +0.055$, one-sided lower 95\% bound $0.045$, $p<10^{-3}$). The first comparison is the practically relevant one for deciding whether the three-model ensemble is worth its additional inference cost: the gain over the best single model is small but statistically reliable. Under the 7--9B parameter scale and the single-GPU constraint considered here, \emph{well-tuned supervised encoders with small ensembles outperform instruction-tuned open LLMs} on this fine-grained, imbalanced task.

\subsection{Prompt sensitivity of the LLM baselines}
\label{sec:prompt-sensitivity}

Instruction-tuned models are sensitive to prompt design, so a comparison against supervised encoders should not rest on a single prompt. We accordingly swept four zero-shot prompt styles (\texttt{direct}, \texttt{cot\_hidden}, \texttt{definition} and \texttt{qa}) across all four models, together with few-shot prompting over exemplar counts $k \in \{1,2,4,8,12,16,20\}$ and the QLoRA variants. Every template is reproduced verbatim in \ref{app:prompts}, and per-configuration results appear in Section~S5 of the Supplementary Material. Table~\ref{tab:prompt-sensitivity} reports the zero-shot sweep on the validation split.

\begin{table*}[t]
\caption{Prompt sensitivity of the instruction-tuned LLM baselines: validation macro $F_1$ over the 19 values for each of the four zero-shot prompt styles. The best style per model is shown in bold. Prompting configurations are selected on validation only; the selected configuration is then evaluated once on test (Table~\ref{tab:transformers-vs-llms}).}
\centering
\small
\setlength{\tabcolsep}{6pt}
\fitwidth{%
\begin{tabular}{l c c c c c}
\toprule
\textbf{Model} & \texttt{direct} & \texttt{cot\_hidden} & \texttt{definition} & \texttt{qa} & \textbf{Range} \\
\midrule
Gemma 2 9B      & 0.184 & 0.182 & \textbf{0.215} & 0.176 & 0.039 \\
Llama 3.1 8B    & 0.119 & 0.100 & \textbf{0.163} & 0.097 & 0.066 \\
Ministral 8B    & 0.106 & 0.087 & \textbf{0.177} & 0.102 & 0.090 \\
Qwen 2.5 7B     & \textbf{0.091} & 0.037 & 0.068 & 0.061 & 0.054 \\
\bottomrule
\end{tabular}}
\label{tab:prompt-sensitivity}
\end{table*}

Prompt style matters, and it matters unevenly. Within a single model, the choice of style moves validation macro-$F_1$ by roughly a factor of two for Ministral~8B ($0.087$--$0.177$) and by almost two and a half for Qwen~2.5~7B ($0.037$--$0.091$), whereas Gemma~2~9B is comparatively stable ($0.176$--$0.215$). The best style is also not the same for every model: \texttt{definition}, which supplies a one-line gloss of each of the 19 values, wins for three of the four models, but \texttt{direct} wins for Qwen~2.5~7B. Supplying explicit value definitions is thus the single most useful prompt-design choice here, while the hidden chain-of-thought style is never the best option and is clearly the worst for Qwen~2.5~7B.

Few-shot prompting adds further gains that grow with the number of exemplars but do not change the ordering. With the \texttt{definition} style, validation macro-$F_1$ rises from $0.204$ at $k=1$ to $0.225$ at $k=20$ for Gemma~2~9B, from $0.141$ to $0.182$ for Llama~3.1~8B, and from $0.085$ to $0.168$ for Ministral~8B.

For each model we selected the configuration that maximised validation macro-$F_1$ and carried only that configuration to the test set, which is why Table~\ref{tab:transformers-vs-llms} lists a single prompt per model. This matters for the interpretation of Section~\ref{sec:results-llms}: the gap between supervised encoders and instruction-tuned LLMs is not an artefact of one untested prompt, because each LLM is represented by its own best validation-selected prompt, the most favourable setting our sweep identifies. Any less carefully chosen prompt would only widen the gap.

\subsection{Summary of findings}
\label{sec:results-summary}

Taken together, the results answer our research questions as follows:

\begin{itemize}[leftmargin=1.5em]
  \item \textbf{RQ1 (presence).} Moral presence is reliably learnable from single sentences: several DeBERTa-based gates achieve $F_{1} \approx 0.73$ on the test set.
  \item \textbf{RQ2 (hierarchy).} Presence-gated hierarchies do not clearly outperform direct multi-label detectors under the same 8\,GB budget; gate recall becomes a bottleneck.
  \item \textbf{RQ3 (calibration and lightweight signals).} Decision-threshold calibration is the single most effective lever, lifting the text-only baseline by about $+0.03$ macro-$F_1$ (from $0.282$ to $0.315$)---more than any individual feature. On top of it, simple auxiliary signals (LIWC-22, short-range context, topics) add at most $+0.005$ macro-$F_1$ on average, a difference that a paired per-seed test does not distinguish from seed variance; we report them as inconclusive rather than as gains.
  \item \textbf{RQ4 (LLMs).} At the 7--9B scale and under the single-8\,GB-GPU setting described in Section~\ref{sec:hardware-software}, instruction-tuned LLMs (zero-/few-shot and QLoRA) and their ensembles remain noticeably behind supervised DeBERTa, while a compact DeBERTa ensemble attains our best macro-$F_1 = 0.332$.
\end{itemize}

Taken together, these findings are less about introducing a novel architectural component and more about clarifying which existing modelling patterns are robust and cost-effective for refined Schwartz value detection at sentence level. By holding the dataset, training protocol, and hardware budget fixed while varying only a few architectural and feature choices, the paper provides actionable guidance on when presence gating is (not) beneficial, which lightweight signals are worth adding, and how medium-sized encoders and 7--9B instruction-tuned LLMs compare under realistic deployment constraints.

\subsection{Discussion and implications}
\label{sec:discussion-implications}

We now discuss how these findings inform sentence-level value detection and the design of value-aware NLP systems.

First, the results for RQ1 show that sentence-level \emph{moral presence} is a meaningful and learnable construct on the refined Schwartz continuum. Presence gates recover the existence of any value with $F_{1} \approx 0.73$ despite sparse and often implicit cues, supporting the ValuesML and ValueEval choice to annotate values at the sentence level \citep{ValueEval24Zenodo,Kiesel2024}. At the same time, the remaining errors reflect that many value cues are only fully interpretable in context, highlighting a tension between sentence-level operationalisations and more distributed, document-level conceptions of values.

Second, the comparison between direct and presence-gated architectures (RQ2) provides negative evidence for a class of pipelines that are often assumed to help moral and value-related NLP. In our setting, both approaches share the same encoder and compute budget, yet presence gating does not yield clear gains over direct prediction. Hard pre-filtering by a gate can discard sentences where values are present but only weakly signalled, especially for rare values. This suggests that joint modelling of presence and specific values within one multi-label classifier may be preferable when the label space is fine-grained and imbalanced.

Third, error patterns along the Schwartz continuum (Section~\ref{sec:error-analysis}) align with the underlying motivational geometry \citep{Schwartz2012}. Many errors are confusions between neighbouring or closely related values (e.g., \emph{Benevolence} vs.\ \emph{Universalism}, \emph{Security: societal} vs.\ \emph{Conformity: rules}, \emph{Power: dominance} vs.\ \emph{Power: resources}), rather than across distant sectors of the circle. This suggests that evaluation schemes which weight errors by distance on the continuum may better reflect the substantive severity of mistakes than flat macro-$F_1$.

A further finding with direct practical implications concerns decision-threshold calibration. Because the 19-value task is severely imbalanced, the default decision threshold $t=0.5$ systematically favours the majority negative class and under-detects rare values. Tuning a global threshold on the validation set recovers a large part of this loss: it improves the text-only baseline by about $+0.03$ macro-$F_1$ (from $0.282$ to $0.315$; Section~\ref{sec:results-lightweight}), more than any single auxiliary feature contributes, and on its own brings a standard baseline to or above the best official ValueEval'24 English run. We therefore treat threshold calibration not as routine preprocessing but as a primary, standalone design recommendation: for imbalanced value or moral-detection tasks, threshold (or label-wise) tuning should be exhausted before investing in additional features, lexica, or ensembling.

What this contributes is a controlled decomposition of where the performance actually comes from. Calibrating a single global decision threshold accounts for roughly two thirds of the total improvement over the best official ValueEval'24 English run, more than every architectural and feature choice we tested combined, and it reproduces on a second encoder family under an identical protocol. A practitioner who adopts the elaborate pipeline but omits the threshold sweep incurs most of the cost and little of the benefit. Our results also contribute to the discussion on encoder models vs.\ instruction tuned LLMs for structured classification. Under the explicit single-8\,GB-GPU constraint and fine-grained, imbalanced value taxonomy considered here, carefully tuned DeBERTa-base models with lightweight features and compact ensembles outperform 7--9B open LLMs in zero-/few-shot and QLoRA setups (RQ4). This does not contradict evidence that very large proprietary models internalise broad moral distinctions, but it nuances expectations that generative LLMs will automatically dominate specialised encoders on all moral or value-related tasks. Several structural properties of the task likely drive the gap at the 7--9B scale: (i) the 19-way, fine-grained label space---with several motivationally adjacent values---exceeds what zero-/few-shot prompting can reliably separate from a handful of in-context examples; (ii) the machine-translated input may diverge from the text distributions these models were primarily trained on; (iii) the severe class imbalance lets a model reach adequate overall accuracy by abstaining from rare values, which depresses macro-$F_1$ even when frequent values are handled well; and (iv) QLoRA fine-tuning on this relatively small training set may be insufficient for a 7--9B model to internalise the fine-grained Schwartz distinctions that a fully fine-tuned encoder acquires directly. We do not disentangle these factors empirically, but naming them clarifies when medium-sized open LLMs are likely to trail specialised encoders on similar fine-grained, imbalanced tasks. For practitioners with limited compute, investing in supervised encoders, threshold calibration, and small ensembles appears more effective than deploying medium-scale LLMs as classifiers.

A further practical question is whether the added complexity is worth its cost. Largely it is not. Threshold calibration is a validation-set sweep over twenty candidate values: it requires no retraining, adds nothing to inference, and yields $+0.033$ macro-$F_1$. The three-model soft-voting ensemble, by contrast, triples both training and inference cost and buys $+0.012$ over its strongest single component (one-sided lower 95\% bound $0.007$, $p<10^{-3}$); it is a real but small gain, and it is the only one of our elaborations that survives statistical scrutiny. The auxiliary features sit below that: they add training-time complexity and external lexicon dependencies for differences we cannot separate from seed noise. Therefore, a practitioner working under a comparable budget should calibrate first, ensemble only if the last percentage point matters and the threefold inference cost is acceptable, and treat feature engineering as the lowest-priority option rather than the first.

From an applied perspective, several design recommendations follow:

\begin{itemize}[leftmargin=1.5em]
  \item \textbf{Calibrate decision thresholds first.} Under class imbalance, tuning the decision threshold on validation recovers more macro-$F_1$ ($\approx\!+0.03$ here) than any single feature we tested, at negligible compute cost; it should be the first step before adding features or ensembling.
  \item A strong presence gate can serve as a cheap filter to identify value-rich sentences for downstream analysis (e.g., qualitative coding or targeted annotation), even if full 19-way predictions remain noisy.
  \item \textbf{Do not expect much from lightweight lexical features.} Once the threshold is calibrated, LIWC-22, moral lexica and topic features added at most $+0.005$ macro-$F_1$ in our experiments, and none of these gains survived a paired per-seed test. They are cheap enough to be worth trying, but they should not be budgeted for as a reliable source of improvement, and a reported gain of this size should be checked against seed variance before it is believed.
  \item Small ensembles of diverse but inexpensive models provide consistent gains over single systems in both the DeBERTa and LLM families and are preferable to relying on a single, more complex architecture.
\end{itemize}

Compared to prior ValueEval systems and broader moral NLP work \citep[e.g.,][]{Kiesel2023,Kiesel2024,Rezapour2021,Trager2022,Reinig2024}, our study systematically compares direct and hierarchical pipelines, lightweight features, and instruction-tuned LLMs under the same dataset and hardware constraints, with explicit statistical testing. Grounding the analysis in the refined Schwartz continuum also highlights how its richer, circular value space shapes both model errors and evaluation.

\subsection{Error analysis and qualitative cases}
\label{sec:error-analysis}

The macro-$F_1$ scores leave substantial room for improvement, especially at the value level (best supervised ensemble $F_{1}{=}0.332$), which motivates a brief error analysis.

\paragraph{Effect of prevalence and sparse labels}
As expected from the strong skew in per-value prevalence (Section~\ref{sec:stats}, \ref{app:prevalence}), models perform markedly better on frequent values than on rare ones. Values such as \emph{Security: societal}, \emph{Conformity: rules}, \emph{Power: dominance}, and \emph{Power: resources} (each present in roughly 4--9\% of sentences) achieve the highest per-label $F_1$ scores in our direct DeBERTa runs, with relatively balanced precision and recall. Extremely rare values such as \emph{Humility}, \emph{Hedonism}, \emph{Self-direction: thought}, or \emph{Universalism: tolerance} often sit near the floor, with $F_1$ dominated by either very low recall or noisy high-recall regimes. Threshold calibration improves the trade-off but does not remove this basic sparsity effect.

\paragraph{Confusions along the continuum}
Per-value behaviour is governed largely, but not entirely, by prevalence. Across the three seeds, per-value test $F_1$ for the calibrated baseline correlates strongly with test-split prevalence (Spearman $\rho = 0.59$, $p = 0.008$ over the 19 values): the eight values occurring in under $2\%$ of sentences average $F_1 = 0.238$, against $0.402$ for the six occurring in over $4\%$. Per-value scores are reported in Section~S2 of the Supplementary Material.

The informative cases are the ones that break this pattern. \emph{Universalism: nature} is rare ($2.0\%$ of test sentences) yet is the best-detected value of all ($F_1 = 0.578$), and \emph{Tradition} is rarer still ($1.4\%$) yet second-best ($0.466$). Both are expressed through highly distinctive vocabulary (environment, climate and conservation in the first case, religion, custom and heritage in the second), so a modest number of positive examples suffices. Conversely, \emph{Self-direction: action} is comparatively frequent ($3.5\%$) but poorly detected ($0.218$), and \emph{Conformity: interpersonal} ($1.3\%$, $0.151$) and \emph{Humility} ($0.2\%$, $0.021$) sit at the floor. These are abstract, diffusely expressed motivations with no characteristic lexicon, and they are frequently realised implicitly rather than stated. The practical implication is that scarcity of positive examples and lexical diffuseness are separate problems: more annotation would plausibly help \emph{Humility}, whereas \emph{Self-direction: action} appears limited by how the value is expressed rather than by how often.

Many errors are plausible confusions between neighbouring values on the Schwartz circle. For instance, sentences that express solidarity with specific groups are sometimes annotated as \emph{Universalism: concern} but predicted as \emph{Benevolence: caring}, or vice versa. Sentences about protecting society, maintaining order, or respecting the law often trigger both \emph{Security: societal} and \emph{Conformity: rules}, while the gold labels may mark only one of them. Similar boundary effects appear between \emph{Power: dominance} and \emph{Power: resources} or between \emph{Self-direction: action} and \emph{Achievement}. In many such cases the model predictions are semantically defensible yet counted as errors because the annotation selects a single dominant value.

We tested this continuum effect quantitatively for the best ensemble. We counted pairwise confusions over the 19 values by treating, for each test sentence, every missed gold value (false negative) together with every spuriously predicted value (false positive) as a substitution, summed over the test set and symmetrised per value pair. Raw confusion frequency is dominated by prevalence---frequent values such as \emph{Security: societal} take part in many confusions irrespective of their position on the circle---so the raw correlation with angular distance is weak (Spearman $\rho = -0.10$, $p = 0.18$). Once prevalence is controlled, by normalising each pair's count by the two values' total error involvement (a confusion-affinity measure), a modest but statistically significant continuum effect emerges: confusion affinity is significantly higher for values that lie \emph{closer} on the Schwartz circle (Spearman $\rho = -0.18$, $p = 0.02$ across the 171 value pairs), and the ten highest-affinity pairs have a mean circular distance of $3.3$ steps versus $5.0$ expected by chance (permutation $p = 0.018$). The highest-affinity pairs are intuitive neighbours---for example \emph{Stimulation} vs.\ \emph{Hedonism}, \emph{Conformity: interpersonal} vs.\ \emph{Humility}, and \emph{Universalism: concern} vs.\ \emph{Universalism: tolerance}---so, after accounting for class frequency, the model's errors do concentrate between motivationally adjacent values rather than across the circle.

Overall, the error analysis confirms three main points: (i) data sparsity remains a primary bottleneck for rare values, even with strong encoders and ensembles; (ii) after controlling for class frequency, mistakes concentrate between neighbouring values on the refined continuum rather than across distant sectors; and (iii) evaluation schemes that account for the circular value geometry are therefore well motivated and may better capture the substantive severity of errors in future work.

\subsection{Effect of machine translation on model performance}
\label{sec:provenance}

Only about 14\% of the English ValueEval'24 release originates from English texts (Section~\ref{sec:dataset}), and the lexical and affective features are word-frequency based, so they are the signals most exposed to translation artefacts. The question that bears on this paper's conclusions is not the intrinsic quality of the translations but whether translation changes what the models do. We can address that directly, because source language is recoverable at no cost: ValueEval'24 \texttt{Text-ID}s carry an ISO language prefix, so every test sentence can be assigned to its original language and the test set stratified accordingly.

We evaluate the calibrated text-only baseline over the three seeds, applying each seed's single global threshold $t^\star$ unchanged to every stratum, exactly as a deployed system would. Table~\ref{tab:provenance} gives the breakdown.

\begin{table*}[t]
\caption{Test macro $F_1$ by original source language, for the calibrated text-only baseline (mean $\pm$ std over seeds 42, 7, 1701; each seed's global $t^\star$ applied unchanged). English is the only stratum that is not machine-translated.}
\centering
\small
\setlength{\tabcolsep}{6pt}
\fitwidth{%
\begin{tabular}{l r c l r c}
\toprule
\textbf{Source} & \textbf{$n$} & \textbf{Macro $F_1$} & \textbf{Source} & \textbf{$n$} & \textbf{Macro $F_1$} \\
\midrule
Hebrew     & 1{,}480 & $0.361 \pm 0.006$ & Bulgarian & 1{,}302 & $0.291 \pm 0.009$ \\
Greek      & 1{,}402 & $0.333 \pm 0.013$ & French    &   931 & $0.264 \pm 0.019$ \\
Turkish    & 2{,}187 & $0.310 \pm 0.009$ & Dutch     & 2{,}150 & $0.262 \pm 0.002$ \\
German     & 1{,}874 & $0.291 \pm 0.003$ & \emph{English (original)} & 2{,}005 & $0.254 \pm 0.014$ \\
           &         &                   & Italian   & 1{,}238 & $0.231 \pm 0.007$ \\
\bottomrule
\end{tabular}}
\label{tab:provenance}
\end{table*}

The aggregate result runs opposite to the concern that motivates the question. Macro-$F_1$ on the English-original stratum is $0.254 \pm 0.014$, whereas on the machine-translated remainder it is $0.320 \pm 0.004$: translated sentences are handled \emph{better}, by $0.066$, and English-original text sits in the lower half of the per-language ordering.

This difference is a corpus-composition artefact rather than a translation effect. The two strata differ sharply in how much value content they contain: $28.3\%$ of English-original sentences carry at least one value, against $54.4\%$ of translated sentences, and the overall positive rate is $1.56\%$ versus $3.29\%$. Macro-$F_1$ over 19 rare labels is strongly prevalence-dependent, so the sparser stratum is intrinsically harder at any fixed threshold. No label has zero positives in either stratum, so the gap is not an artefact of degenerate per-label scores.

Controlling for this removes the effect. Matching the translated stratum to the English-original one on the per-sentence value-count distribution and resampling $1{,}000$ times, the gap collapses and changes sign: the prevalence-matched translated macro-$F_1$ is $+0.001$, $+0.011$ and $+0.032$ \emph{below} the English-original value for seeds 42, 7 and 1701 respectively, and English-original falls inside the matched 95\% bootstrap interval for two of the three seeds. Therefore, we find no evidence that machine translation degrades value detection on this corpus. The honest statement is the negative one: the controlled difference is of the same order as its own seed-to-seed variation, so the finding is an absence of detectable degradation, not a benefit of translation.

We can also test the more specific hypothesis that word-frequency features are the exposed signal. Adding LIWC-22 to the calibrated baseline improves macro-$F_1$ by $+0.010$ on English-original sentences and by $+0.004$ on translated ones. The direction is consistent with the hypothesis: lexicon features do appear to help more where the text was not translated. The difference of $0.006$ is nevertheless within the seed-to-seed variation of the English stratum ($\approx 0.010$--$0.014$), so we report it as suggestive rather than established.

Two limits of this analysis should be stated. First, it measures whether translation affects \emph{model performance}, which is the question relevant to our conclusions; it is not a translation-quality audit, and a direct study of the English release's translation quality remains outstanding (Section~\ref{sec:limitations-ethics}). Second, source language is confounded with outlet, genre, topic and annotation team, and matching on value count does not control for those; the per-language spread in Table~\ref{tab:provenance} is wide enough ($0.231$ to $0.361$) that some of it is likely attributable to such factors rather than to translation.

\subsection{Limitations and ethical considerations}
\label{sec:limitations-ethics}

Our study has several limitations that should be kept in mind when interpreting the results. First, we work only with the English, machine-translated release of the ValueEval'24 corpus. We do not use the original-language texts, and we do not model potential translation artefacts or cultural differences in how values are expressed. This means that our findings may not directly transfer to other languages or domains, even within the ValuesML collection. Second, we operationalise human values at the sentence level and treat sentences as independent instances. Many value cues, however, are distributed across longer stretches of discourse; our models do not exploit document-level structure, speaker-level profiles, or longitudinal information. Third, the task is constrained by the size and composition of the dataset: it covers news and political manifestos on contemporary policy topics, with a highly imbalanced 19-value label space and some degree of annotation subjectivity that we cannot fully quantify.

Machine translation is, moreover, not neutral with respect to our auxiliary signals, and it affects most of our data: only about 14\% of the release is English-original (Section~\ref{sec:dataset}), the remainder being machine-translated from eight other languages. The lexicon-based and affective features---LIWC-22, the eMFD, the Moral Justification Dictionary, and the NRC VAD, EmoLex, and Emotion-Intensity lexica together with WorryWords---are computed from surface word frequencies and are therefore the components most exposed to translation artefacts: machine translation can replace value-laden terms with near-synonyms that fall outside a dictionary's entries, render hedging either too literally or too smoothly, and flatten culturally specific ways of expressing values, biasing these features in directions that are hard to detect from English-only evaluation. Topic features are derived from the same translated text, whereas the contextual DeBERTa encoder is presumably more robust to individual lexical substitutions. To our knowledge, no study has specifically quantified the machine-translation quality of the ValueEval'24 English release or its downstream effect on value detection; assessing this---for example by comparing original-language and machine-translated lexical features, or through a targeted manual audit of translation quality---is an important priority for future work.

A further limitation concerns model capacity and hardware. All experiments are run under a single 8\,GB GPU constraint, which restricts us to DeBERTa-base encoders and 7--9B open instruction-tuned LLMs with parameter-efficient fine-tuning. Larger models and more expensive architectures may perform differently, and our conclusions are therefore best read as guidance for settings with similar compute budgets. In particular, because our evidence rests on a single corpus, a single encoder family (DeBERTa-base), and one LLM scale band (7--9B, 4-bit), the extent to which these architectural comparisons transfer to other corpora, encoders, or model scales remains an open empirical question rather than a demonstrated property. Finally, we use a specific combination of auxiliary features (short-range context, LIWC-22, moral lexica, topics) and simple voting-based ensembles. Other feature sets or aggregation schemes might yield additional gains, but exploring them systematically is beyond the scope of this work.

Because we analyse value-related content in political and news texts, ethical considerations are important. The ValueEval'24 data are released under a restricted Data Usage Agreement and based on previously collected documents; we follow the license by releasing only our code, configurations, tuned thresholds, and per-model predictions, but not the texts themselves. Our models are intended for aggregate analyses of corpora (e.g., to study value patterns in political discourse), not for profiling individuals or supporting automated decision-making about persons or groups. Predicted values should not be treated as ground-truth moral profiles, and any use for targeting, ranking, or labelling individuals would raise serious ethical and legal concerns. More broadly, Schwartz values are a theory-driven construct; model outputs necessarily reflect this framework and its cultural assumptions and should be interpreted with caution, especially in cross-cultural comparisons.

\section{Conclusions and future work}
\label{sec:conclusions-future-work}

We have studied sentence-level detection of human values on the refined Schwartz continuum using the English, machine-translated ValueEval'24 corpus. Each sentence is annotated with 19 values and a derived \texttt{presence} label indicating whether any value is active. Within this setting, we formulated two tasks---binary detection of moral presence and multi-label prediction of the 19 values---and evaluated a range of architectures under an explicit single 8\,GB GPU constraint.

Our experiments yield four main conclusions. First, moral presence is a learnable signal at the sentence level: several DeBERTa-based presence models reach positive-class $F_1 \approx 0.73$ despite subtle cues and class imbalance. Second, under matched compute and careful threshold calibration, a presence-gated hierarchy does not clearly outperform a direct multi-label DeBERTa classifier on the 19 values; in practice, gate recall becomes a bottleneck and direct architectures remain competitive or superior. Third, once the decision threshold is calibrated, lightweight auxiliary signals such as short-range context, LIWC-22 and moral lexica, and topic features add at most $+0.005$ macro-$F_1$ on average, a difference a paired per-seed test cannot separate from seed variance. Nearly all of the improvement over a text-only baseline comes from calibration rather than from the added features. Fourth, at the explored 7--9B scale and with QLoRA fine-tuning, instruction-tuned open LLMs lag behind well-tuned DeBERTa-base models and their small ensembles on this fine-grained, imbalanced value taxonomy.

From a practical perspective, these results suggest that, for sentence-level value detection under realistic compute budgets, medium-sized encoders with calibrated thresholds and compact ensembles are a strong and cost-effective choice. A simple presence classifier can already support workflows that require filtering value-rich sentences, while a direct multi-label DeBERTa ensemble enriched with lightweight features offers the best trade-off between performance and hardware requirements in our experiments.

Several directions follow from this work. One natural extension is to move beyond independent sentences and incorporate richer discourse context, for example by modelling paragraphs or full documents and aggregating value signals across sentences. A second direction is to exploit the geometry of the Schwartz continuum more directly, for instance by using loss functions or evaluation measures that weight confusions by their angular distance on the circle rather than treating all errors equally. Third, addressing the extreme imbalance of rare values will require data-centric strategies such as targeted annotation, active learning, or semi-supervised methods that leverage unlabeled political and news text. Finally, as larger and more capable LLMs become available under feasible hardware constraints, it will be important to revisit the comparison between supervised encoders and generative models, and to explore hybrid systems in which LLMs assist with annotation, explanation, or interactive analysis rather than replacing specialised classifiers.

Taken together, our findings provide empirical evidence and practical guidance for building value-aware NLP models on the refined Schwartz continuum under limited compute. We hope that this work will support future studies of political and media discourse, help to clarify the strengths and limits of sentence-level value detection, and motivate further research on robust, interpretable, and socially responsible value-aware NLP.

\section{CRediT authorship contribution statement}
\label{sec:credit-authorship}

\ifanon
Available upon publication of this paper.
\else
\textbf{Víctor Yeste:} Conceptualization, Methodology, Software, Validation, Formal analysis, Investigation, Resources, Data Curation, Writing - Original Draft, Writing - Review \& Editing, Visualization, Project administration. \textbf{Paolo Rosso:} Supervision, Writing - Review \& Editing.
\fi

\section{Funding}
\label{sec:funding}

This research did not receive any specific grant from funding agencies in the public, commercial, or not-for-profit sectors.

\section{Declaration of generative AI and AI-assisted technologies in the manuscript preparation process}
\label{sec:declaration-generative-ai}

During the preparation of this work, the authors used GPT-5.1 Thinking \citep{ChatGPT2026} and Claude Opus 5 \citep{ClaudeOpus2026} to improve the readability and language of the manuscript, to assist with drafting text, and to help implement the evaluation and analysis scripts released with this work. The study design, the choice of experiments, the interpretation of results, and the conclusions drawn are the authors' own. The authors verified every reported figure against the released code and outputs, reviewed and edited the content as needed, and take full responsibility for the content of the published article.

\section{Data availability}
\label{sec:data-availability}

This study uses the English, machine-translated release of the ValueEval'24 dataset provided by the shared-task organizers \citep{ValueEval24Zenodo,Kiesel2024}. The corpus is distributed under a restricted Data Usage Agreement that permits research use but prohibits redistribution of the texts. Researchers can obtain the original data, including train/validation/test splits, by registering and downloading the official release from Zenodo \citep{ValueEval24Zenodo}.

To support reproducibility while respecting this license, we release only derived artefacts, not the texts themselves. Specifically, we provide:

\ifanon
\begin{itemize}
  \item \textbf{Code and configurations:} all training, evaluation, threshold-tuning, and ensembling scripts, together with configuration files specifying model architectures, hyperparameters, and random seeds, at GitHub.\footnote{URL available upon publication of this paper.}
  \item \textbf{Tuned thresholds and predictions:} per-model decision thresholds selected on the validation set, and per-model and ensemble predictions for the official ValueEval'24 English splits.
  \item \textbf{Environment and documentation:} a description of the software environment and step-by-step instructions for reproducing the experiments assuming access to the official ValueEval'24 data.
  \item \textbf{Model checkpoints:} the best performing fine-tuned DeBERTa models are released on Hugging Face.\footnote{URL available upon publication of this paper.}. These checkpoints are trained exclusively on the official ValueEval'24 training split and do not contain any text from the corpus.
\end{itemize}
\else
\begin{itemize}
  \item \textbf{Code and configurations:} all training, evaluation, threshold-tuning, and ensembling scripts, together with configuration files specifying model architectures, hyperparameters, and random seeds, at GitHub.\footnote{\url{https://github.com/VictorMYeste/human-value-detection}}
  \item \textbf{Tuned thresholds and predictions:} per-model decision thresholds selected on the validation set, and per-model and ensemble predictions for the official ValueEval'24 English splits.
  \item \textbf{Environment and documentation:} a description of the software environment and step-by-step instructions for reproducing the experiments assuming access to the official ValueEval'24 data.
  \item \textbf{Model checkpoints:} the best performing fine-tuned DeBERTa models are released on Hugging Face.\footnote{\url{https://huggingface.co/papers/2601.14172}}. These checkpoints are trained exclusively on the official ValueEval'24 training split and do not contain any text from the corpus.
\end{itemize}
\fi

Given access to the dataset under its original terms, these artefacts are sufficient to reproduce all results reported in this paper.

\appendix
\setcounter{table}{0}

\section{Per-value prevalence by split}
\label{app:prevalence}

Percentages are computed over the number of sentences in each split (Train $=$ 44{,}758; Validation $=$ 14{,}904; Test $=$ 14{,}569). Counts in parentheses are rounded to the nearest integer. These detailed prevalences complement the split-level summary in Table~\ref{tab:corpus-stats} and the imbalance discussion in Section \ref{sec:stats}.

\begin{table*}
\caption{Per-value prevalence by split (\% of sentences; raw counts in parentheses).}
\centering
\small
\fitwidth{%
\begin{tabular}{lccc}
\toprule
\textbf{Value} & \textbf{Train} & \textbf{Validation} & \textbf{Test} \\
\midrule
Self-direction: thought         & 1.29\% (577)  & 1.15\% (171) & 1.17\% (170) \\
Self-direction: action          & 3.61\% (1{,}616) & 3.26\% (486) & 3.51\% (511) \\
Stimulation                     & 2.62\% (1{,}173) & 2.82\% (420) & 2.55\% (372) \\
Hedonism                        & 0.86\% (385)  & 0.67\% (100) & 0.86\% (125) \\
Achievement                     & 6.42\% (2{,}873) & 6.37\% (949) & 6.25\% (911) \\
Power: dominance                & 4.63\% (2{,}072) & 4.40\% (656) & 4.33\% (631) \\
Power: resources                & 5.00\% (2{,}238) & 4.86\% (724) & 5.53\% (806) \\
Face                            & 1.81\% (810)  & 1.90\% (283) & 1.83\% (267) \\
Security: personal              & 2.03\% (909)  & 1.87\% (279) & 2.42\% (353) \\
Security: societal              & 8.95\% (4{,}006) & 8.46\% (1{,}261) & 7.90\% (1{,}151) \\
Tradition                       & 1.20\% (537)  & 1.84\% (274) & 1.35\% (197) \\
Conformity: rules               & 6.10\% (2{,}730) & 6.41\% (955) & 6.25\% (911) \\
Conformity: interpersonal       & 1.35\% (604)  & 1.37\% (204) & 1.34\% (195) \\
Humility                        & 0.24\% (107)  & 0.29\% (43)  & 0.21\% (31)  \\
Benevolence: caring             & 2.29\% (1{,}025) & 2.29\% (341) & 2.22\% (323) \\
Benevolence: dependability      & 1.94\% (868)  & 1.93\% (288) & 1.98\% (288) \\
Universalism: concern           & 4.97\% (2{,}224) & 4.50\% (671) & 5.04\% (734) \\
Universalism: nature            & 2.05\% (918)  & 2.57\% (383) & 2.01\% (293) \\
Universalism: tolerance         & 1.07\% (479)  & 0.81\% (121) & 1.17\% (170) \\
\midrule
\textit{Any value present (presence)} & 51.53\% (23{,}064) & 50.99\% (7{,}600) & 50.81\% (7{,}403) \\
\bottomrule
\end{tabular}}
\label{tab:prevalence-by-split}
\end{table*}

\section{LLM prompt templates and decoding configuration}
\label{app:prompts}

This appendix reproduces, verbatim, the prompts and inference settings used for every LLM configuration reported in Section~\ref{sec:llms} and in the Supplementary Material. They are transcribed from the released implementation rather than reconstructed. Where a template interpolates a long constant, the constant is given once below and referenced as \texttt{<VALUES>} or \texttt{<DEFINITIONS>}; \texttt{\{sentence\}} is replaced by the sentence being classified. Lines too long for the text width are wrapped with a two-space continuation indent, which is not present in the underlying strings.

\paragraph{System prompt}
The same system message is used for all models and all prompt styles. Where a chat template does not support a system role, it is prepended to the user message instead.

\begin{small}\begin{verbatim}
You are a moral-psychology assistant. Using the “refined basic
  values” taxonomy (Schwartz 1992; Schwartz et al. 2012), answer
  the user’s labeling requests exactly as instructed.
\end{verbatim}\end{small}

\paragraph{\texttt{<VALUES>}}
The 19 refined values, comma-separated, in the fixed order used throughout.

\begin{small}\begin{verbatim}
Self-direction: thought, Self-direction: action, Stimulation,
  Hedonism, Achievement, Power: dominance, Power: resources, Face,
  Security: personal, Security: societal, Tradition, Conformity:
  rules, Conformity: interpersonal, Humility, Benevolence: caring,
  Benevolence: dependability, Universalism: concern, Universalism:
  nature, Universalism: tolerance
\end{verbatim}\end{small}

\paragraph{\texttt{<DEFINITIONS>}}
One-line glosses of the 19 values, after \citet{Schwartz2012}. Used only by the \texttt{definition} style.

\begin{small}\begin{verbatim}
- **Self-direction: thought**: Freedom to cultivate one’s own
  ideas and abilities
- **Self-direction: action**: Freedom to determine one’s own
  actions
- **Stimulation**: Excitement, novelty, and change
- **Hedonism**: Pleasure and sensuous gratification
- **Achievement**: Success according to social standards
- **Power: dominance**: Power through exercising control over
  people
- **Power: resources**: Power through control of material and
  social resources
- **Face**: Maintaining one’s public image and avoiding
  humiliation
- **Security: personal**: Safety in one’s immediate environment
- **Security: societal**: Safety and stability in the wider
  society
- **Tradition**: Maintaining and preserving cultural, family, or
  religious traditions
- **Conformity: rules**: Compliance with rules, laws, and formal
  obligations
- **Conformity: interpersonal**: Avoidance of upsetting or harming
  other people
- **Humility**: Recognising one’s insignificance in the larger
  scheme of things
- **Benevolence: caring**: Devotion to the welfare of in-group
  members
- **Benevolence: dependability**: Being a reliable and trustworthy
  member of the in-group
- **Universalism: concern**: Commitment to equality, justice, and
  protection for all people
- **Universalism: nature**: Preservation of the natural
  environment
- **Universalism: tolerance**: Acceptance and understanding of
  those who are different from oneself
\end{verbatim}\end{small}

\paragraph{Zero-shot templates}
The four styles compared in Section~\ref{sec:prompt-sensitivity}.

\begin{small}\begin{verbatim}
[direct]
List **only** the basic values that are clearly expressed or implied
in the SENTENCE.
Return them as a JSON array of strings (choose from: <VALUES>). No
other text.

SENTENCE: {sentence}

[cot_hidden]
Think step-by-step about which of the 19 basic human values are
present in the SENTENCE, **but do not reveal your reasoning**.
Output **only** the JSON array of value names.

Allowed values: <VALUES>

SENTENCE: {sentence}

[definition]
### Value definitions
<DEFINITIONS>

### Task
Identify which of the above values the SENTENCE relates to.  Return
**only** a JSON array of the matching value names.

SENTENCE: {sentence}

[qa]
Question: Which basic human values are present in the SENTENCE?
Answer: Return **only** a JSON array (strings) chosen from:
<VALUES>

SENTENCE: {sentence}
\end{verbatim}\end{small}

\paragraph{Few-shot construction}
For $k>0$, $k$ exemplars are rendered with the template below and concatenated with newlines, and the block is prepended to the zero-shot prompt of the selected style. Exemplar labels are serialised as a JSON array. The separator on the third line is an en-dash triple.

\begin{small}\begin{verbatim}
SENTENCE: {example_sentence}
OUTPUT: {example_labels}
–––
\end{verbatim}\end{small}

\paragraph{Decoding and output parsing}
Generation is greedy in every configuration: sampling is disabled, so \emph{temperature}, \emph{top-$p$} and \emph{top-$k$} do not apply. The generation length cap is \texttt{max\_new\_tokens=200}. Models are loaded in 4-bit NF4 via \texttt{bitsandbytes}, as described in Section~\ref{sec:llms}. Generated text is parsed by extracting the first JSON array; recognised value names are matched against the 19-value vocabulary and mapped to a binary 19-dimensional vector, and unparseable or empty generations yield the all-zero vector. Decision thresholds for the resulting scores are tuned on validation and applied unchanged to test, exactly as for the supervised models.

\bibliographystyle{elsarticle-num-names}
\bibliography{references.bib}

\fi 

\end{document}